# Advanced Gesture Recognition for Autism Spectrum Disorder Detection: Integrating YOLOv7, Video Augmentation, and VideoMAE for Naturalistic Video Analysis

Amit Kumar Singh[0009–0000–3900–4015] and Vrijendra Singh[0000–0002–8818–5673]

Department of Information Technology,
Indian Institute of Information Technology, Allahabad, 211013, India
amitramraj005@gmail.com , vrij@iiita.ac.in

**Abstract.** Deep learning and contactless sensing technologies have significantly advanced the automated assessment of human behaviors in healthcare. In the context of autism spectrum disorder (ASD), repetitive motor behaviors such as spinning, head banging, and arm flapping are key indicators for diagnosis. This study focuses on distinguishing between children with ASD and typically developed (TD) peers by analyzing videos captured in natural, uncontrolled environments. Using the publicly available Self-Stimulatory Behavior Dataset (SSBD), we address the classification task as a binary problem—ASD vs. TD—based on stereotypical repetitive gestures. We adopt a pipeline integrating YOLOv7-based detection, extensive video augmentations, and the VideoMAE framework, which efficiently captures both spatial and temporal features through a high-ratio masking and reconstruction strategy. Our proposed approach achieves **95% accuracy**, **0.93 precision**, **0.94 recall**, and **0.94 F1 score**, surpassing the previous state-of-the-art by a significant margin. These results demonstrate the effectiveness of combining advanced object detection, robust data augmentation, and masked autoencoder-based video modeling for reliable ASD vs. TD classification in naturalistic settings.

## 1 INTRODUCTION

Repetitive behavioral patterns and difficulty with communication and social interaction are among the many symptoms together referred to as autism spectrum disorder (ASD) [1]. According to the CDC's Autistic and Developmental Disabilities Monitoring (ADDM) Network, about 1 in 44 American children received an ASD diagnosis in 2020 [2]. In 2018, Dr. Narendra Arora [3] carried out a research across many Indian areas and discovered that one in every eight children has at least one neurodevelopmental problem and one in every 100 people under the age of ten has autism. Opinions from pediatricians and parents are essential for the early diagnosis of ASD. New study, however, indicates that many children do not receive a conclusive diagnosis for autism until they are much older (beyond the age of three) [4]. It's still challenging to identify and diagnose ASD in children before the age of two.

Numerous behavioral and motor characteristics, including stereotypically repetitive motions, can be used to diagnose autism. Such repeated motions are usually the result of passion, anxiety, or fury. Among the signs of autism are erratic behaviors that occur unexpectedly and for short bursts of time. Most of the time, the youngster hurts himself with these motions. Sadly, parents could fail to see these motions and fail to notice them. It is essential to identify these repeated actions in order to diagnose autism in youngsters. Psychologists must spend time and money monitoring their behavior, and the financial burden falls on the family. Recent advancements in technology and deep learning algorithms have made it feasible for artificial intelligence devices to automatically recognize self-stimulatory movements.

The aim of this study is to develop a model capable of recognizing children's repetitious behavior. This would assist doctors in selecting the most appropriate behavioral therapy and provide family support in remote places where access to state-of-the-art diagnostic technologies is limited.

With the use of contemporary technologies like computer vision and deep learning models, it may be possible to identify children's stereotypical repetitive motions in order to provide a precise diagnosis and prompt therapy intervention. This work developed a real-time computer vision and deep learning system to monitor and identify archetypal repetitive behavior in youngsters.

- We designed a novel model by leveraging the VideoMAE model for video classification, which combines masked autoencoders for feature extraction and attention mechanisms for gesture classification, and trained this model for the detection of stereotypical repetitive autistic behaviors.
- Our trained VideoMAE model performs significantly better than prior vision-based techniques, according to experimental results on the Self-Stimulatory Behavior Dataset (SSBD) [5]. This provides a solid baseline for the detection of stereotypical repetitive autistic actions in videos.

## 2 RELATED WORK

Various approaches have been developed by researchers to examine and monitor behavior to identify autism, such as questionnaires [6–10], eye gaze [11–14], functional magnetic resonance imaging (fMRI) [15–19], facial feature analysis [20–22] and multiple modalities in a single study [23–25]. Most of these studies aim to identify and classify ASD and typical-developed children with different modalities of data. In addition, stereotypical repetitive autistic behaviors have been used to evaluate autism, with most researchers using the SSBD dataset [5]. To evaluate autism with autistic behavior firstly, it is important to classify these stereotypical repetitive behaviors.

In the early years, the researcher used haar cascade for person detection and classified the gestures using basic ML or DL algorithms such as MLP, SVM, DCNN, ConvLSTM, etc., and achieved an accuracy of 79% [26]. The SSBD was proposed by Rajagopalan et al. [5,27] and comprises movies showing autistic children going about their regular lives. They combined a histogram of optical flow with a histogram of dominating movements. Their three-way challenge headbanging, spinning, and hand-flapping classification model produced an accuracy of 76.3%. Single CNN was used to extract the features from each input sequence of headbanging videos and fed into LSTM for further classification. Deng A et al. [28] introduced the video swin transformer for analyzing autistic behaviors and combined visual features with language information and achieved an accuracy of 97%. Washington et al. [29] presented a new model and achieved the mean F1-score of 90.77%. Anish et al. [30] implemented LSTM and MobileNetV2 on SSBD for abnormal hand movement classification, achieving an F1 score of 75.2 ± 0.6. Wei et al. [31] Employed MS-TCN on a modified SSBD dataset, achieving 84% weighted F1 score. They addressed the noise problem in the dataset and resolved it by replacing 11 videos from the actual dataset. Dia et al. [32] Utilized the Perceive model on SSBD and Affect-Net datasets, focusing on facial expressions in ASD children, with 74.5% accuracy. An approach for the behavioral diagnosis of ASD was created by Ali et al. [33]. During the time of their ASD diagnosis, they gathered and documented a collection of recordings of stereotypical children's behavior captured in an uncontrolled setting. The dataset contained 388 videos of 5 categories and combined of the two streams of Inflated 3D Networks (I3D) produced the greatest accuracy (85.6 to 86.04%). Singh and Singh [34] presented a three-class classification approach for autism-related gesture recognition using the Self-Stimulatory Behavior Dataset (SSBD). Their framework combined YOLOv7-based detection, video augmentation, and the VideoMAE architecture to identify repetitive behaviors such as spinning, head banging, and arm flapping in naturalistic videos. While the study reported an accuracy of 97.7%, surpassing prior methods, its focus was limited to classifying specific repetitive behaviors rather than broader ASD vs. TD classification.

However, these state-of-the-art works recognize stereotypical repetitive autistic behaviors in autistic children on the SSBD dataset are not very optimized (less than 78%). We also find that the publicly available dataset has a very small amount of data with a high amount of noise due to being collected from an uncontrolled environment. The initial research work did not acquire features very accurately due to noisy data.

To address these issues, we have implemented the VideoMAE model, which combines masked autoencoders for improved feature extraction and attention mechanisms to optimize the classification performance.

# 3 MATERIAL AND METHODS

Among children with ASD, stereotypical repeated movements are one of the most often seen stimming activities. Three different types of repeated gestures are taken into consideration in this study: arm flipping, head banging, and spinning. These behaviors are characterized by recurrent actions that may only continue for a moment. There are several ways in which arm flapping might appear as a stimming behavior: jerking fingers, clicking fingers, and massive arm movements.

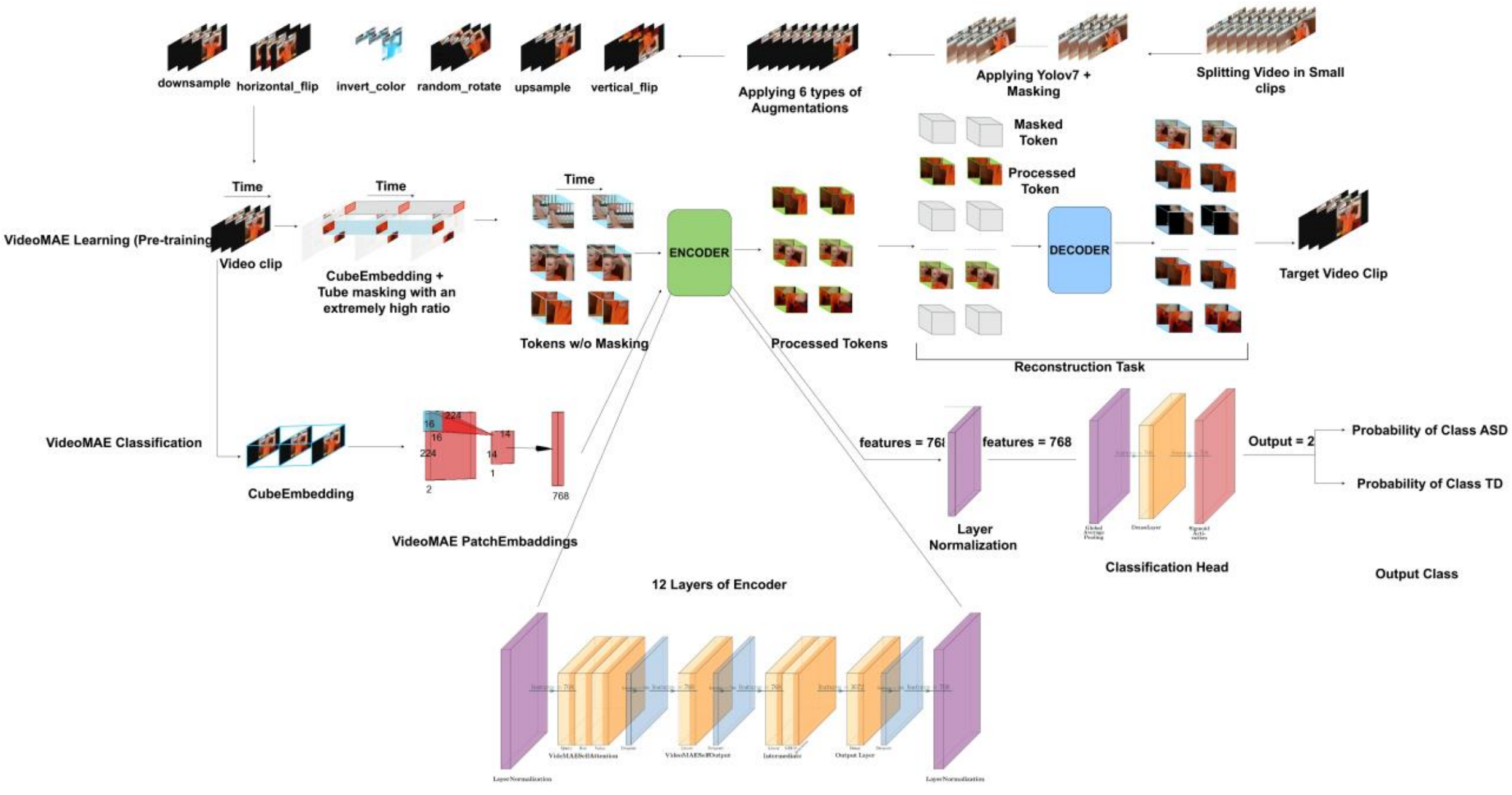

**Fig.1.** Work Flow of VideoMAE based model

Children with autism are susceptible to headbanging, a kind of stimming activity in which they repeatedly strike their head against something like a floor, furniture, door, etc. This might lead to self-harm. Another stimming habit is spinning, in which a child spins things or themselves. Children with autism often utilize these behaviors to control sensory input or to deal with stress or excitement. Obtaining a video dataset is the first stage in our work.The second stage involves trimming the SSBD videos, where the videos are segmented into smaller, more manageable clips. In the third stage, YOLOv7 with masking is applied to these trimmed videos, allowing for precise object detection and masking within each segment. The fourth stage focuses on applying video augmentations, which involves employing various augmentation techniques to increase the diversity and robustness of the video data. These stages collectively prepare the data for further analysis and model training, ultimately supporting the project's goals of effective behavior analysis.(as illustrated in Fig 1).

## 3.1 SSBD DATASET

The study utilizes Self-Stimulatory Behavioural Data (SSBD) [5], which is specifically designed to aid in the identification and analysis of stereotypical repetitive behaviors in autistic children. It's a comprehensive dataset comprising video recordings of children exhibiting self-stimulatory stimming behaviors of three different kinds: arm flapping, head banging, and spinning (Fig 2). These "stimming" behaviors are characterized by repetitive or unusual movements or noises, often observed in individuals with Autism Spectrum Disorders (ASD). Parents or caregivers captured these videos in unmanaged natural environments.

The SSBD dataset contains a total of 75 videos. Each category includes 25 videos. However, some of the videos are not available at the given URL, due to YouTube privacy concerns. We were able to retrieve only 59 videos: 19 for Spinning, 20 for Head Banging, and 19 for Arm Flapping.

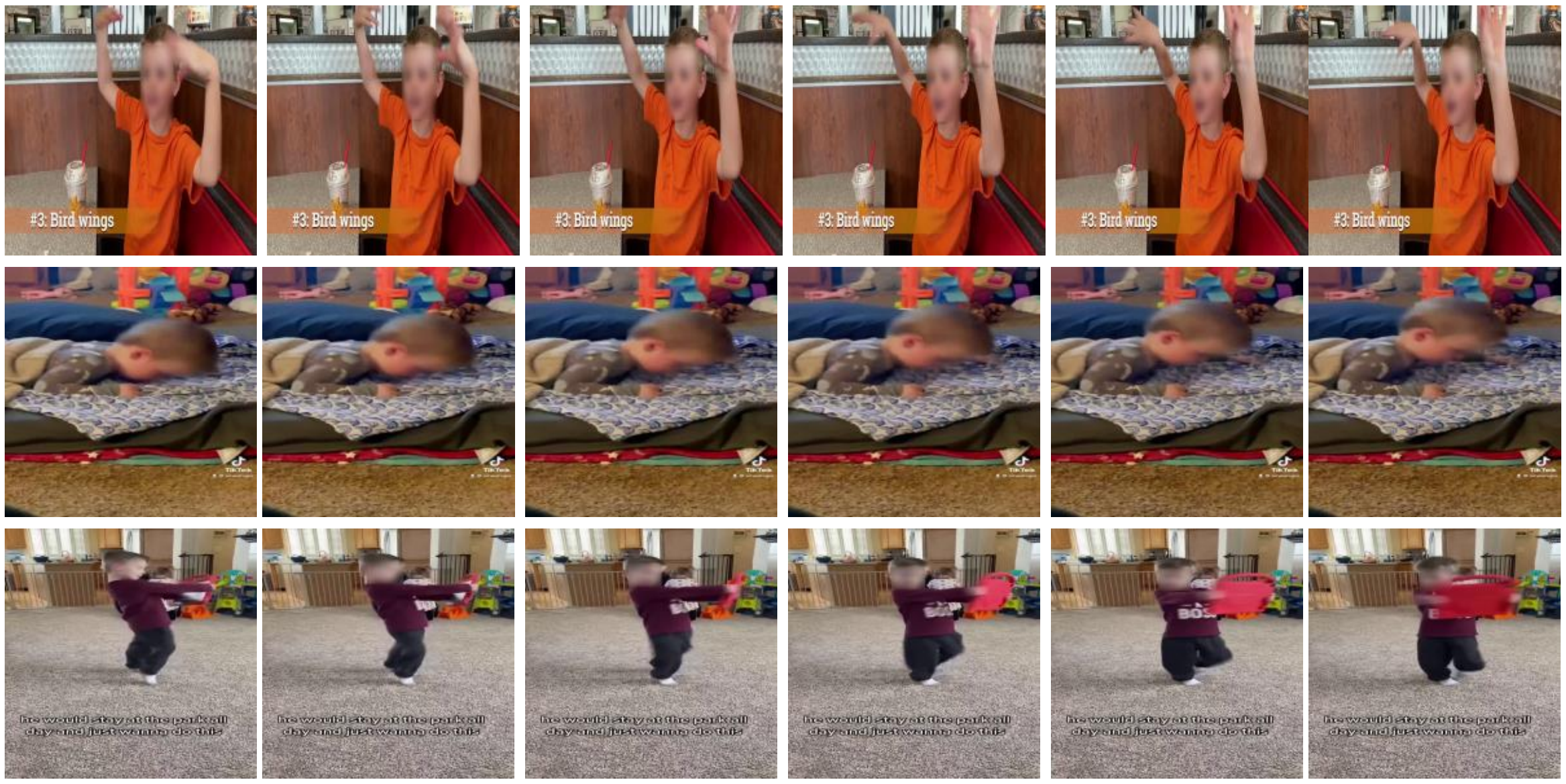

**Fig.2.** Each class includes a frame depicting a certain action. The first row illustrates arm flapping, the second row illustrates head pounding, and the last row illustrates spinning motion [5]

## 3.2 DATA PRE-PROCESSING

The initial step in our data processing pipeline involves trimming the videos. The SSBD provided the URLs of 75 YouTube videos, with annotations at the start and finish of repetitive actions. Every video has an average length of 90 seconds.

**Table 1.** Number of videos in each class before and after pre-processing

| | SSBD | | | Modified SSBD | | | SSBD Plus | | |
|---|---|---|---|---|---|---|---|---|---|
| | original | trim | augmented | original | trim | augmented | original | trim | augmented |
| ArmFlapping | 25 | 29 | 203 | 18 | 43 | 301 | 34 | 42 | 294 |
| HeadBanging | 25 | 41 | 287 | 20 | 41 | 287 | 7 | 15 | 105 |
| Spinning | 25 | 54 | 378 | 20 | 63 | 435 | 16 | 36 | 252 |
| TD (New Class) [44] | 36 | 36 | 252 | 36 | 36 | 252 | 36 | 36 | 252 |

The table 1 provides a detailed overview of the number of videos in each class at various stages of preprocessing across three datasets: SSBD [5], Modified SSBD [35], and SSBD Plus [36]. Each dataset undergoes three stages: original, trimmed, and augmented.

**SSBD** : Initially, each class—ArmFlapping, HeadBanging, and Spinning—has 25 videos. After trimming, the counts increase to 29, 41, and 54, respectively. Post-augmentation, these numbers rise significantly to 203, 287, and 378.

**Modified SSBD** : In the original dataset, the number of videos for ArmFlapping, HeadBanging, and Spinning starts at 18, 20, and 20, respectively. Trimming increases these numbers to 43, 41, and 63, and augmentation further expands them to 301, 287, and 435.

**SSBD Plus** : ArmFlapping begins with 34 videos, trimmed to 42, and grows to 294 after augmentation. HeadBanging starts with 7 videos, increases to 15 after trimming, and reaches 105 post-augmentation. Spinning starts with 16 videos, increases to 36, and ends with 252 after augmentation. Additionally, the new class **”TD” (Typical Developed)** remains consistent across all datasets and pre-processing stages, with 36 original videos, 36 trimmed, and 252 augmented.

This table 2 provides a comprehensive statistical analysis of the SSBD dataset and its variations, trim - SSBD and trim SSBD Aug, focusing on behaviors such as Arm Flapping, Head Banging, and Spinning.

**Table 2.** Detailed statistics description for the SSBD dataset, including minimum and maximum frame counts, frame resolutions, sizes, and average video duration across different behaviors like Arm Flapping, Head Banging, and Spinning, both for the original and trimmed datasets

| | SSBD | | | trim SSBD | | | trim SSBD Aug | | |
|---|---|---|---|---|---|---|---|---|---|
| | ArmFlapping | HeadBanging | Spinning | ArmFlapping | HeadBanging | Spinning | ArmFlapping | HeadBanging | Spinning |
| Min Frame Count | 44 | 31 | 90 | 44 | 14 | 55 | 22 | 7 | 28 |
| Max Frame Count | 713 | 720 | 1656 | 258 | 227 | 270 | 258 | 227 | 270 |
| Avg Frame Count | 191.04 | 330.68 | 407.36 | 135.65 | 121.29 | 147.03 | 125.99 | 112.65 | 136.56 |
| Min Frame Size | 204X360 | 176X144 | 204X360 | 204X360 | 176X144 | 270X360 | 204X360 | 176X144 | 270X360 |
| Max Frame Size | 1280X720 | 1280X720 | 1280X720 | 1280X720 | 1280X720 | 1280X720 | 1920X1080 | 1920X1080 | 1920X1080 |
| Avg Frame Size | 644.88X486.72 | 513.04X436.48 | 580.16X565.76 | 644.20X477.51 | 540.24X460.29 | 702.40X602.81 | 690.21X511.62 | 578.82X493.17 | 752.57X645.87 |
| Avg video duration | 7.08 sec | 11.79 sec | 17.22 sec | 5.04 sec | 4.41 sec | 6.16 sec | 4.68 sec | 4.10 sec | 5.72 sec |
| Number of videos | 25 | 25 | 25 | 29 | 41 | 50 | 203 | 287 | 378 |

It outlines several key metrics including frame counts, frame sizes, video durations, and the number of videos. The original SSBD dataset features a minimum frame count of 44 for Arm Flapping and a maximum of 1656 for Spinning, while the trimmed dataset reduces these ranges to 44 and 270 frames respectively, showing the impact of data trimming. The augmented dataset further modifies these counts, with a minimum frame count as low as 22 for Arm Flapping and a maximum of 270, demonstrating the augmentation’s flexibility. Average frame counts reveal typical video lengths, with Spinning having the longest average at 407.36 frames in the original dataset.

Frame size metrics show the variability in video quality, ranging from a minimum resolution of 176x144 pixels in Head Banging videos to a maximum of 1920x1080 pixels in the augmented dataset. The average frame size also reflects this diversity, with the original dataset showing resolutions like 644.88x486.72 pixels for Arm Flapping, increasing to 752.57x645.87 pixels for Spinning in the augmented version. Average video durations, which are 7.08 seconds for Arm Flapping and 17.22 seconds for Spinning in the original dataset, decrease in the trimmed and augmented versions, indicating shorter clips due to editing and augmentation strategies. The number of videos also sees a substantial increase in the augmented dataset, from 25 in the original to 378 for Spinning, highlighting extensive augmentation efforts aimed at enhancing dataset diversity and size. This detailed overview is essential for understanding the dataset’s structure and its application in behavior analysis, providing valuable insights for research and modeling purposes.

**Table 3.** Detailed statistics description for the Modified SSBD dataset, including minimum and maximum frame counts, frame resolutions, sizes, and average video duration across different behaviors like Arm Flapping, Head Banging, and Spinning, both for the original and trimmed datasets

| | modified SSBD | | | trim modified SSBD | | | trim modified SSBD Aug | | |
|---|---|---|---|---|---|---|---|---|---|
| | ArmFlapping | HeadBanging | Spinning | ArmFlapping | HeadBanging | Spinning | ArmFlapping | HeadBanging | Spinning |
| Min Frame Count | 246 | 88 | 190 | 70 | 45 | 61 | 35 | 11 | 3 |
| Max Frame Count | 7813 | 11913 | 4452 | 382 | 430 | 235 | 382 | 430 | 222 |
| Avg Frame Count | 2403.38 | 2122.4 | 1890.95 | 165.44 | 185.56 | 164.17 | 145.60 | 145.46 | 136.56 |
| Min Frame Size | 204X360 | 192X144 | 202X360 | 204X360 | 192X144 | 202X360 | 204X360 | 192X144 | 202X360 |
| Max Frame Size | 1280X720 | 1280X720 | 1280X720 | 1280X720 | 1280X720 | 1280X720 | 1920X1080 | 1920X1080 | 1920X1080 |
| Avg Frame Size | 640.11X449.33 | 655.8X466.4 | 500.9X396.0 | 530.51X395.72 | 583.6X426.73 | 568.31X411.42 | 568.39X423.98 | 625.31X457.21 | 615.19X441.70 |
| Avg video duration | 92.04 sec | 74.13 sec | 77.89 sec | 6.66 sec | 6.77sec | 6.16 sec | 5.93 sec | 5.29 sec | 6.14 sec |
| Number of videos | 18 | 20 | 20 | 43 | 41 | 63 | 301 | 287 | 435 |

This table 3 presents a detailed statistical analysis of the Modified SSBD dataset and its variations trim modified SSBD and trim modified SSBD Aug—across three behaviors: Arm Flapping, Head Banging, and Spinning. It outlines critical metrics such as minimum and maximum frame counts, average frame sizes, video durations, and the number of videos. The minimum frame counts for the original dataset are 246 for Arm Flapping and 190 for Spinning, while the trimmed dataset shows significantly lower minimums, dropping to 70 for Arm Flapping and just 3 for Spinning. Maximum frame counts reveal that the original Arm Flapping videos can reach up to 7813 frames, whereas the trimmed version caps at 430 frames for Head Banging.

Average frame counts indicate a substantial reduction in video lengths, with Arm Flapping averaging 2403.38 frames in the original dataset compared to 165.44 frames in the trimmed version. Frame sizes show variability as well, with minimum resolutions starting at 192x144 pixels for Head Banging and maximum sizes of 1280x720 pixels in both the original and trimmed datasets. The augmented dataset further improves resolution, reaching up to 1920x1080 pixels. Average video durations are particularly notable, with Arm Flapping lasting an average of 92.04 seconds in the original dataset, significantly reduced to 6.66 seconds in the trimmed version.

The number of videos per behavior also highlights the extensive augmentation efforts; Arm Flapping increases to 301 videos in the augmented dataset, demonstrating a commitment to enhancing dataset diversity and size. This comprehensive statistical overview is essential for understanding the dataset's structure and its applicability in behavior analysis research, offering valuable insights for future studies in this domain.

**Table 4.** Detailed statistics description for the SSBD-Plus dataset, including minimum and maximum frame counts, frame resolutions, sizes, and average video duration across different behaviors like Arm Flapping, Head Banging, and Spinning, both for the original and trimmed datasets

| | SSBD plus | | | trim SSBD plus | | | trim SSBD plus Aug | | |
|---|---|---|---|---|---|---|---|---|---|
| | ArmFlapping | HeadBanging | Spinning | ArmFlapping | HeadBanging | Spinning | ArmFlapping | HeadBanging | Spinning |
| Min Frame Count | 31 | 49 | 121 | 30 | 49 | 60 | 11 | 25 | 30 |
| Max Frame Count | 450 | 1032 | 1277 | 203 | 192 | 211 | 201 | 192 | 210 |
| Avg Frame Count | 144.38 | 302.28 | 382.93 | 102.5 | 116.33 | 158.88 | 93.31 | 106.60 | 144.59 |
| Min Frame Size | 202X360 | 202X360 | 184X360 | 202X360 | 202X360 | 184X360 | 202X360 | 202X360 | 184X360 |
| Max Frame Size | 1280X720 | 1280X720 | 1280X720 | 1280X720 | 1280X720 | 1280X720 | 1920X1080 | 1920X1080 | 1920X1080 |
| Avg Frame Size | 622.64X660 | 760.5X617.14 | 616.5X653.12 | 587.33X637.14 | 462.66X480 | 610.88X573.05 | 629.23X682.65 | 495.61X514.28 | 654.52X613.96 |
| Avg video duration | 10.34 sec | 11.51 sec | 11.27 sec | 3.68 sec | 4.55 sec | 5.51 sec | 3.35 sec | 4.17 sec | 5.01 sec |
| Number of videos | 34 | 7 | 15 | 42 | 15 | 36 | 294 | 105 | 252 |

This table 4 provides a detailed statistical analysis of the SSBD-Plus dataset and its variations, including trim SSBD plus and trim SSBD plus Aug, across three behaviors: Arm Flapping, Head Banging, and Spinning. It outlines essential metrics such as minimum and maximum frame counts, average frame sizes, video durations, and the number of videos. The minimum frame counts show that Arm Flapping videos start at 31 frames, while Spinning has a minimum of 121 frames in the original dataset. The trimmed dataset reflects minor reductions, with minimum counts of 30 for Arm Flapping and 60 for Spinning. Maximum frame counts vary significantly, with Arm Flapping reaching up to 450 frames and Spinning up to 1277 frames in the original dataset.

Average frame counts indicate a notable decrease in video lengths in the trimmed versions, with Arm Flapping averaging 102.5 frames, compared to 144.38 frames in the original dataset. Frame sizes show consistency in minimum resolutions at 202x360 pixels across most behaviors, while maximum sizes reach 1280x720 pixels for the original dataset, and the augmented version improves resolution to 1920x1080 pixels. Average frame sizes also vary, with Arm Flapping averaging 622.64x660 pixels in the original dataset, while the trimmed version averages 587.33x637.14 pixels.

Video durations further highlight the differences, with the original dataset averaging 10.34 seconds for Arm Flapping, significantly reduced to 3.68 seconds in the trimmed dataset. The number of videos increases dramatically in the augmented dataset, with Arm Flapping reaching 294 videos, showcasing extensive augmentation efforts to enhance dataset diversity and size. This comprehensive overview of the SSBD-Plus dataset is crucial for understanding its structure and relevance for behavior analysis research, providing insights into its potential applications in the field.

**Table 5.** Dataset Details

| | SSBD | | | Modified SSBD | | | Typical Developed |
|---|---|---|---|---|---|---|---|
| | ArmFlapping | HeadBanging | Spinning | ArmFlapping | HeadBanging | Spinning | |
| Min Frame Count | 44 | 31 | 90 | 246 | 88 | 190 | 30 |
| Max Frame Count | 713 | 720 | 1656 | 7813 | 11913 | 4452 | 633 |
| Avg Frame Count | 191.04 | 330.68 | 407.36 | 2403.38 | 2122.4 | 1890.95 | 192.05 |
| Min Frame Size | 204X360 | 176X144 | 204X360 | 204X360 | 192X144 | 202X360 | 320X240 |
| Max Frame Size | 1280X720 | 1280X720 | 1280X720 | 1280X720 | 1280X720 | 1280X720 | 320X240 |
| Avg Frame Size | 644.88X486.72 | 513.04X436.48 | 580.16X565.76 | 640.11X449.33 | 655.8X466.4 | 500.9X396.0 | 320X320 |
| Avg video duration | 7.08 sec | 11.79 sec | 17.22 sec | 92.04 sec | 74.13 sec | 77.89 sec | 7.51 sec |

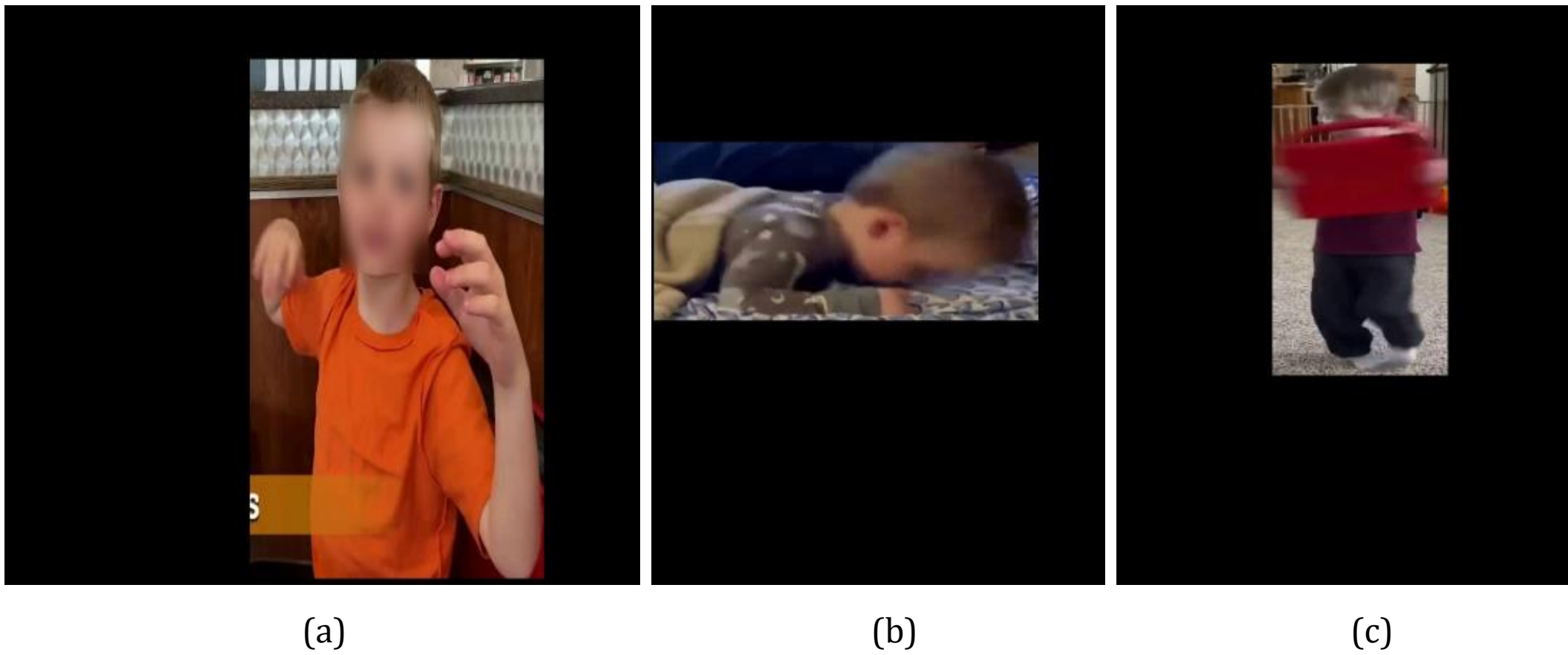

**Fig.3.** Frame from each class, with the first figure showing the arm flapping, the second figure showing the head banging and the last figure showing the spinning behavior after applying Yolov7 and masking. [5]

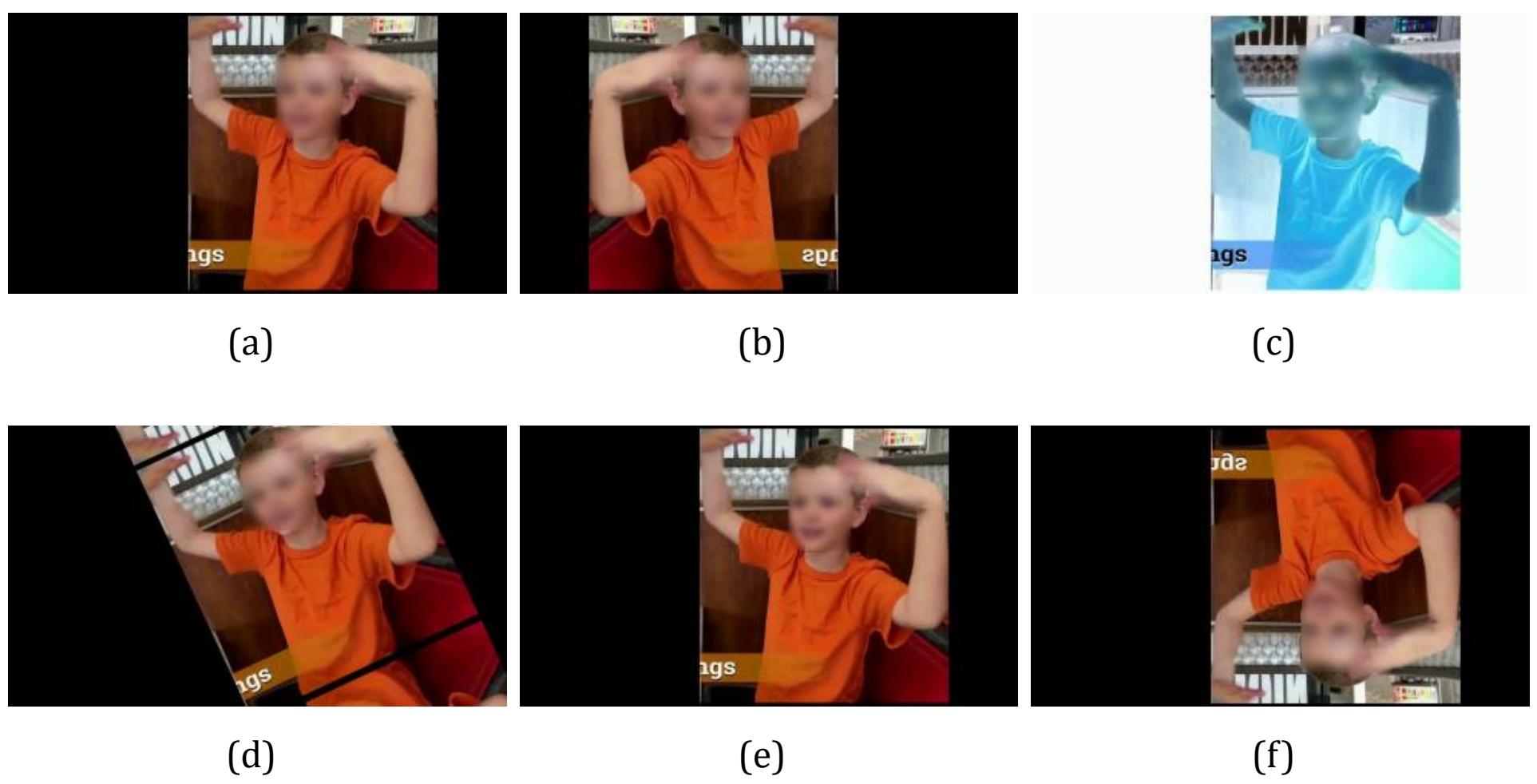

**Fig.4.** Frames showing the augmentations used after applying Yolov7 and masking. (a) downsample, (b) horizontal flip, (c) Invert color, (d) random rotate, (e) upsample, (f) vertical flip

Figure 3 presents representative frames from each behavioral class in the SSBD dataset—arm flapping, head banging, and spinning—after applying Yolov7 detection and masking, which isolates the child and removes background distractions. Figures 4, 5, and 6 illustrate the augmentation strategies used on these frames, including downsampling, upsampling, horizontal and vertical flips, random rotation, and color inversion. Although applied to different samples, the same augmentation pipeline is consistently demonstrated, thereby increasing dataset variability. These preprocessing and augmentation steps are essential for training the VideoMAE model, as they enhance robustness, reduce overfitting, and enable the transformer to learn invariant motion patterns from the relatively small and imbalanced SSBD dataset.

Videos of the SSBD dataset showcase a variety of behavioral movements from various periods. As a result, it was necessary to trim the parts according to behavioral motions. However, due to YouTube privacy issues, only 58 videos are currently available. To maintain the characteristics of the dataset we added 16 new videos of the same duration and same behavior class as missing videos.

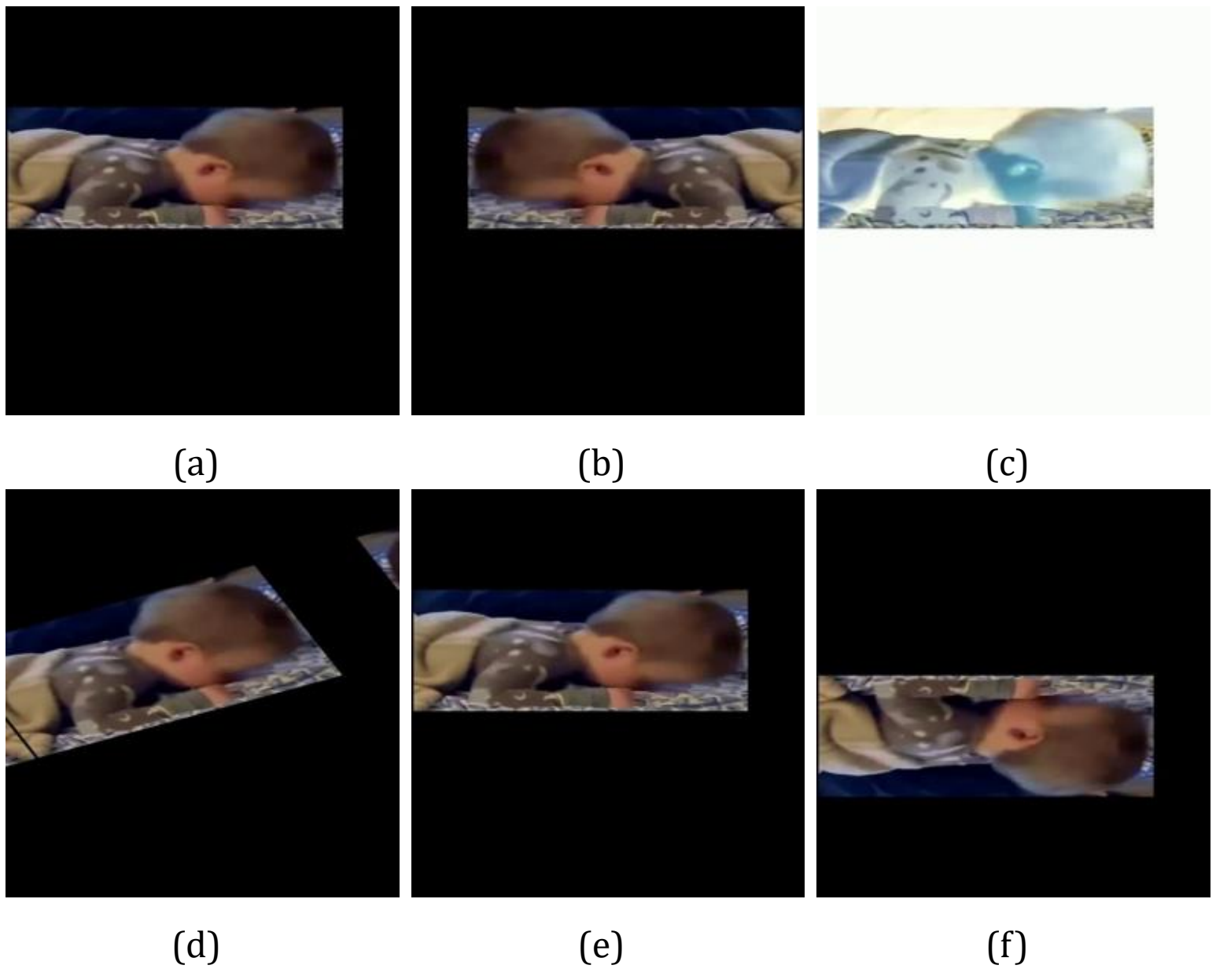

**Fig.5.** Frames showing the augmentations used after applying Yolov7 and masking. (a) downsample, (b) horizontal flip, (c) Invert color, (d) random rotate, (e) upsample, (f) vertical flip

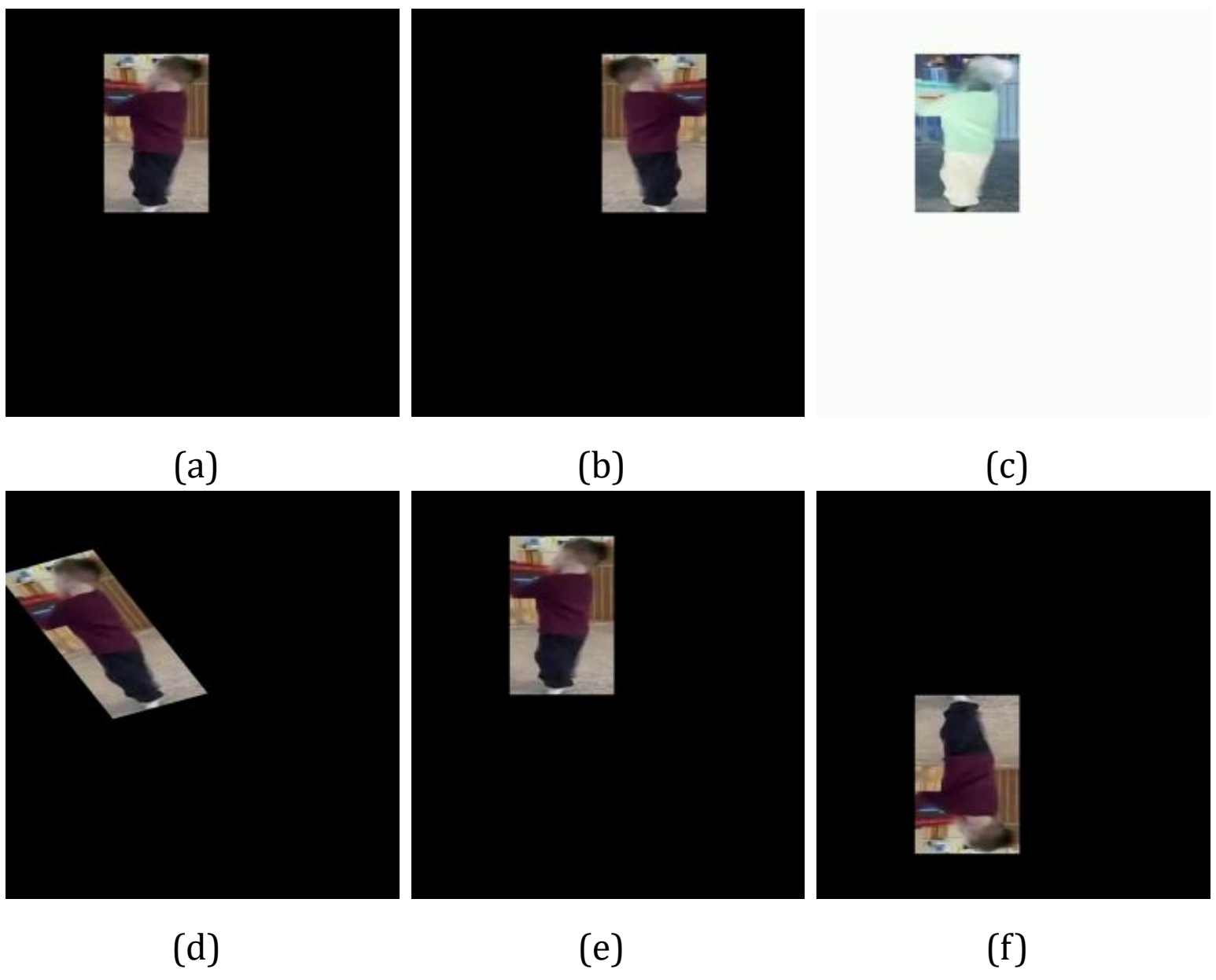

**Fig.6.** Frames showing the augmentations used after applying Yolov7 and masking. (a) downsample, (b) horizontal flip, (c) Invert color, (d) random rotate, (e) upsample, (f) vertical flip

All the Videos were captured in an uncontrolled environment, so there are different types of issues identified in the dataset. For example, too many people in a single frame or only a particular part of the video contain repetitive activity. The dataset was divided into 15% for testing, 15% for validation and 70% for training.

As we can see from the table 2 in the original SSBD dataset, few video are way lengthy then the other video, and these lengthy video have have stimming behaviour for very small duration. Since the dataset was already challenging because of being recorded in natural environment and poor light conditions, we decided to break the long video into smaller parts to ease the training and testing process. Total duration of video in all 3 class remains the same before and after trimming. We are calling this trimmed data as trim SSBD.
After trimming the longer videos in smalller videos the number of videos in each class is shown in the table 2.

After trimming the videos into smaller one, we can see in table 2 that after trimming the videos the overall video duration dropped from 12 sec/video to 5.32 sec/video.

From each video, we took out all frames, which we then preprocessed as follows. All the videos are collected from different resources, so each video has a different dimension. The average height and width of videos is 635 x 526. Since the VideoMAE model requires an input size of 224 x 224, we resized the frames accordingly to fit this requirement.

The table 5 presents a comprehensive overview of three datasets utilized for video analysis: SSBD, Modified SSBD, and Typical Developed. Each dataset focuses on specific behaviors, namely Arm Flapping, Head Banging, and Spinning. The SSBD dataset captures these behaviors with a minimum frame count of 44 for Arm Flapping, 31 for Head Banging, and 90 for Spinning. The maximum frame count in the SSBD dataset ranges from 713 for Arm Flapping to 1656 for Spinning, while the average frame counts are 191.04 for Arm Flapping, 330.68 for Head Banging, and 407.36 for Spinning.

In comparison, the Modified SSBD dataset, which is an enhanced version of the SSBD dataset, shows a significant increase in frame counts across all behaviors. The minimum frame counts in the Modified SSBD dataset are 246 for Arm Flapping, 88 for Head Banging, and 190 for Spinning. The maximum frame counts are substantially higher, with 7813 for Arm Flapping, 11913 for Head Banging, and 4452 for Spinning. Correspondingly, the average frame counts are 2403.38 for Arm Flapping, 2122.4 for Head Banging, and 1890.95 for Spinning.

The Typical Developed dataset, which serves as a baseline for typical behavior development, shows relatively lower frame counts compared to the Modified SSBD dataset. The minimum frame count is 30, the maximum frame count is 633, and the average frame count is 192.05.

The table also details the frame sizes for each dataset. The SSBD dataset has a range of frame sizes, with the minimum sizes being 204x360 for Arm Flapping, 176x144 for Head Banging, and 204x360 for Spinning. The maximum frame size for all behaviors in the SSBD dataset is 1280x720. The average frame sizes are 644.88x486.72 for Arm Flapping, 513.04x436.48 for Head Banging, and 580.16x565.76 for Spinning.

For the Modified SSBD dataset, the minimum frame sizes are consistent with the SSBD dataset at 204x360 for Arm Flapping, 192x144 for Head Banging, and 202x360 for Spinning. The maximum frame size remains at 1280x720 for all behaviors. The average frame sizes are 640.11x449.33 for Arm Flapping, 655.8x466.4 for Head Banging, and 500.9x396.0 for Spinning.

The Typical Developed dataset, however, has a uniform frame size of 320x240 for minimum, maximum, and average values, suggesting a more consistent video resolution.

Additionally, the average video duration varies significantly across the datasets. The SSBD dataset has an average video duration of 7.08 seconds for Arm Flapping, 11.79 seconds for Head Banging, and 17.22 seconds for Spinning. In contrast, the Modified SSBD dataset features much longer videos, with average durations of 92.04 seconds for Arm Flapping, 74.13 seconds for Head Banging, and 77.89 seconds for Spinning. The Typical Developed dataset has an average video duration of 7.51 seconds, which is comparable to the SSBD dataset but much shorter than the Modified SSBD dataset.

### 3.2.1 YOLOv7 [37] Architecture Overview

YOLOv7 (You Only Look Once version 7) is a state-of-the-art deep learning model for real-time object detection. The architecture of YOLOv7 builds upon the innovations of previous YOLO versions, incorporating advancements to enhance accuracy, speed, and robustness. YOLOv7 maintains the fundamental design principles of the YOLO family while integrating several novel techniques to improve performance.

**Fundamental Architecture**

YOLOv7, like its predecessors, is an end-to-end single neural network architecture that predicts bounding boxes and class probabilities directly from full images in one evaluation. This design contrasts with other detection systems where a preselected set of regions is processed. The architecture of YOLOv7 is designed to be fast and efficient, suitable for real-time applications.

**Backbone: CSPDarknet**

The backbone of YOLOv7 is based on a modified version of the CSPDarknet53, a network known for its balance between efficiency and accuracy. The CSP (Cross-Stage Partial Network) structure reduces the redundancy in feature map computation, enhancing the learning capability and the inference speed. The CSPDarknet in YOLOv7 has been adjusted to optimize computational efficiency and accommodate newer hardware capabilities.

**Neck: Path Aggregation Network (PAN)**

Following the backbone, the network uses a Path Aggregation Network (PAN) that enhances the feature hierarchy by aggregating different scales through bottom-up and top-down pathways. This ensures rich semantic information at various resolution scales, which is crucial for detecting objects of different sizes more effectively.

**Head: Detection Layers**

The detection head of YOLOv7 utilizes anchor boxes predefined at different scales to predict bounding boxes relative to these anchors. For each bounding box, the network predicts the center coordinates, dimensions, objectness score (the probability that a box contains an object), and class probabilities. The head processes multiple scales simultaneously, allowing for robust detection across a diverse set of object sizes.

**Enhancements in YOLOv7**

1. **Automatic Bounding Box Clustering**: YOLOv7 automates the selection of anchor box sizes using k-means clustering on the training dataset, improving the alignment between the model's predictions and the typical object dimensions found in the specific data.

2. **EIoU (Eliminate Intersection Over Union)**: YOLOv7 introduces the EIoU loss, which refines the traditional Intersection Over Union (IoU) loss used in bounding box regression. This new metric helps reduce the localization errors in object detection.

3. **Mosaic and MixUp Data Augmentation**: YOLOv7 employs advanced data augmentation strategies like Mosaic and MixUp to enhance the diversity of training examples and improve model generalization. Mosaic combines four training images into a single one, while MixUp blends two images linearly.

4. **CmBN (Cross Mini-Batch Normalization)**: The architecture uses a novel batch normalization technique, CmBN, which stabilizes training in smaller batch sizes by normalizing across multiple minibatches.

5. **SPP (Spatial Pyramid Pooling)**: The inclusion of SPP layers allows the network to maintain spatial hierarchies, providing robustness against object scale variations.

**Training and Inference**

Training YOLOv7 involves using stochastic gradient descent (SGD) or Adam optimization with a specific focus on fine-tuning learning rates, weight decay, and other hyperparameters for optimal convergence. During inference, YOLOv7 employs non-maximum suppression (NMS) to refine the detection bounding boxes, ensuring that each detected object is represented by the most accurate bounding box.

YOLOv7's architecture represents a significant advancement in the field of real-time object detection. It successfully balances speed, accuracy, and the ability to run on various hardware platforms, making it suitable for a wide range of applications from autonomous vehicles to surveillance systems. Its comprehensive design improvements over prior models make it a formidable choice in the competitive landscape of object detection algorithms.

**Handling Multiple Detections in YOLOv7**

In object detection tasks, especially in crowded scenes, it is common for the detection model to identify multiple bounding boxes around objects of interest. In the context of using YOLOv7 for object detection, my approach prioritizes the largest object in view, based on the area of the detected bounding boxes. This strategy is particularly useful in scenarios where the size of the object correlates with its importance or relevance to the task.

**Selection of the Largest Bounding Box**

During the detection phase, after the model predicts bounding boxes for objects in a frame, the following steps are implemented to select the largest box:

1. **Detection and Area Calculation**: For each frame processed by YOLOv7, all detected bounding boxes along with their class probabilities and objectness scores are evaluated. The area of each bounding box is calculated using the formula:**Area** = $(x2 - x1) \times (y2 - y1)$ where x1, y1 and x2, y2 are the coordinates of the top-left and bottom-right corners of the bounding box, respectively.
2. **Identifying the Maximum Area**: Among all detected boxes, the one with the maximum area is selected. This selection is based on the assumption that larger objects are of higher priority:

**Algorithm 1** Selecting the Largest Detected Bounding Box

1: $max_area \leftarrow 0$
2: $max_box \leftarrow None$
3: **for** each detection *d* in detections **do**
4: *xyxy* ← *d.bbox* ▷ Bounding box coordinates
5: *conf* ← *d.conf* ▷ Confidence score
6: *cls* ← *d.cls* ▷ Class identifier
7: *area* ← (*xyxy*[2] − *xyxy*[0]) × (*xyxy*[3] − *xyxy*[1]) ▷ Calculate area
8: **if** *area* > *max area* **then**
9: $max_area \leftarrow area$
10: $max_box \leftarrow xyxy$
11: **end if**
12: **end for**

3. **Cropping the Largest Box**: Once the largest bounding box is identified, it is cropped from the frame. This cropped region can then be processed further or used as the output, depending on the specific requirements of the application.

**Application and Implications**

This methodology is particularly beneficial in applications where the focus is on identifying and tracking the most significant object in a scene, such as in surveillance systems where tracking the largest visible person might be crucial, or in traffic monitoring where larger vehicles might require more immediate attention.

**Benefits**

**Simplicity and Efficiency**: This approach is computationally efficient as it does not require complex postprocessing of all detected objects.

**Focus on Relevant Objects**: By focusing on the largest object, the model ensures that smaller, potentially less relevant detections do not distract from the primary object of interest.

**Limitations**

**Oversight of Smaller Objects**: This method may overlook smaller, yet important, objects in the scene, which might be crucial depending on the application.

**Dependence on Scene Composition**: The effectiveness of this method may vary dramatically with different scene compositions or object distributions.

### Detailed Explanation of Masking in Object Detection

#### Introduction to Masking

In computer vision, masking refers to the process of isolating a specific part of an image or video frame to focus processing on that area. In the context of object detection using the YOLOv7 model, masking is used to highlight or process only the detected objects, specifically the largest object detected in the video frame as per the defined criteria.

### Implementation of Masking

After detecting objects in each frame, the algorithm isolates the largest object detected by applying a mask. This is achieved through the following steps:

1. **Initialization**: A mask is created with the same dimensions as the input frame but contains only zeros. This mask is essentially a binary image where the pixels of interest (i.e., those that belong to the largest detected object) will be set to a high value (255 in this case, representing white), and all other pixels will remain at zero (black).
2. **Drawing the Rectangle**: Once the bounding box with the maximum area is determined (i.e., the largest object), a rectangle is drawn on the mask. The coordinates of this rectangle correspond to the bounding box of the largest object. The rectangle is filled completely (denoted by -1 in the thickness parameter), turning the area inside the bounding box white.

**3.Applying the Mask**: The mask is then applied to the original frame to create a masked image. This step involves a bit-wise operation where the mask is used to keep the pixels within the bounding box unchanged, while all other pixels are turned to black. This highlights the largest detected object by masking out the rest of the frame.

**4.Resizing**

**Purpose of Resizing** Image resizing is a common preprocessing step in computer vision workflows, particularly in scenarios where uniform image sizes are required for further processing, visualization, or storage. Resizing the masked image standardizes the output, ensuring that each frame in the processed video maintains a consistent size, which is crucial for consistent visualization and could be beneficial for any subsequent analysis or machine learning tasks.

**Implementation of Resizing** After the masking step, where the largest detected object in each frame is highlighted, the masked image is resized to specific dimensions. This is implemented by the VideoMAE model. Here is the detailed breakdown of this step:

**Input**: masked image is the image obtained from the previous masking step, where the region outside the largest detected bounding box is blackened, focusing attention on the object.

**Target Size**: The target dimensions for the resizing are specified as (224, 224). This means the output image will be a square of 224 pixels by 224 pixels. Choosing a square shape can be particularly advantageous when dealing with data that will be input into neural networks that expect a fixed size and aspect ratio, or for creating uniformity in video frames for display purposes.

**Interpolation Method**: The **'interpolation=cv2.INTER AREA'** parameter is critical for resizing. Interpolation methods determine how the pixel values are adjusted during resizing:

**INTER AREA**: This method is used for shrinking an image. It may be the preferred interpolation method when the goal is to reduce the resolution of an image while preserving the original appearance. It works by using pixel area relation, making it particularly suitable for making images smaller, as it can help avoid moir´e patterns and maintain image quality.

### Benefits of Resizing in Object Detection Enhancement of Image Details

Resizing the masked images to a larger size, specifically to 720x720 pixels, serves a strategic purpose in scenarios where enhancing the visibility of details within the detected object is crucial. This approach is particularly valuable when the subsequent analysis requires a more granular examination of the object's features, which might be essential for accurate classification, detailed inspection, or precise monitoring tasks.
**Explanation**:
**Increased Resolution**: By enlarging the images, the approach leverages the higher pixel density to better capture and display finer details of the detected object. This is crucial when the object of interest contains features that are critical for further identification or analysis but might be lost at lower resolutions.

**Enhanced Visualization**: Larger images provide a clearer view when visually inspecting the details within the object. This can be especially beneficial for presentations or when manual verification steps are involved in the workflow.

**Improved Analytical Accuracy**: In applications involving detailed feature recognition, such as defect detection in manufacturing or intricate pattern recognition in wildlife monitoring, larger image sizes can allow algorithms to more accurately detect and classify subtle nuances.

**Trade-offs**: While resizing images to a larger size can enhance detail, it is important to acknowledge the trade-offs:

**Increased Computational Load**: Contrary to resizing for reduced dimensionality, increasing image size can lead to a higher computational burden. This requires more memory and processing power, which might affect the throughput of real-time systems.

**Potential Overfitting**: In machine learning scenarios, higher resolution images might lead to models focusing too much on minute, possibly irrelevant details, which can cause overfitting. It's essential to balance the level of detail with the generalizability of the model.

**5. Output**: The processed (masked and resized) image is then written to the output video file, ready for review or further analysis.

**Benefits of Masking in Object Detection**
**Focus on Relevant Objects**: Masking allows for the isolation and focus on significant objects within a scene. In applications such as surveillance, traffic monitoring, or advanced research, focusing on the largest object could be critical for behavior analysis or event detection.
**Reduction of Noise**: By masking out irrelevant parts of the frame, the algorithm reduces the computational load for any subsequent processing and minimizes the noise that could interfere with the analysis. **Enhanced Visualization**: Masking provides a clear visual representation of the detected object, which can be crucial for presentations, further manual assessments, or when visual outputs are necessary for decisionmaking processes.

**Video Augmentation Techniques applied on dataset 1. Horizontal Flip**
**Purpose**: Mirrors each frame of the video across the vertical axis. This augmentation helps in training models to recognize objects irrespective of their orientation, enhancing the robustness of object detection models.
**Implementation**: Utilizes va.HorizontalFlip() from the vidaug library, which applies a horizontal flip transformation to the video frames.
For a frame represented as a matrix F, the horizontal flip can be represented as:
**F _flipped(i, j) = F(i, W - j + 1)**
Where W is the width of the frame,i ranges over the rows, and j ranges over the columns from 1 to W.
**2. Vertical Flip**
**Purpose**: Mirrors each frame across the horizontal axis. Similar to horizontal flipping, vertical flipping helps models learn invariant features, useful in scenarios where vertical orientation can vary, such as aerial imagery analysis.
**Implementation**: Executed through va.VerticalFlip(), flipping the frames upside down.
For the vertical flip, the transformation can be represented as:
**F _flipped(i, j) = F(H - i + 1, j)**
Where H is the height of the frame, i ranges over the rows from 1 to H, and j ranges over the columns.
**3. Upsample**
**Purpose**: Increases the resolution of each frame by a specified scale factor. This is particularly useful for examining finer details within frames, which might be necessary for high-resolution applications or to test the performance of models at higher resolution levels.

**Implementation**: Each frame's dimensions are increased by a factor (e.g., 1.5 times the original size) using cv2.resize with cv2.INTER CUBIC interpolation, which helps in preserving the smoothness of the image.

When upscaling a frame by a factor $\alpha$, the new dimensions are $\alpha$ original dimensions. If F has dimensions W X H, the upsample frame F' has dimensions:

$$W' = \alpha \times W, \qquad H' = \alpha \times H$$

Interpolation is used to estimate pixel values in F', typically using cubic interpolation:

$$F'(i,j) = \text{cubic interpolation}\left(F, \frac{i}{\alpha}, \frac{j}{\alpha}\right)$$

**4. Random Rotate**

**Purpose**: Rotates each frame by a fixed angle (e.g., 25 degrees). Rotation augments are critical for training models to detect objects at various angles, enhancing detection accuracy in uncontrolled environments. **Implementation**: The custom random rotate function uses OpenCV's **cv2.getRotationMatrix2D** and **cv2.warpAffine**, with the rotation center set to the frame's center and **cv2.BORDER REFLECT101** to manage edge pixels.

For a rotation by angle $\theta$ , the transformation matrix M can be given by: $M = \begin{bmatrix} \cos\theta & -\sin\theta & 0 \\ \sin\theta & \cos\theta & 0 \\ 0 & 0 & 1 \end{bmatrix}$

A point (x,y) in the original frame is mapped to: $\begin{bmatrix} x' \\ y' \\ 1 \end{bmatrix} = M \begin{bmatrix} x \\ y \\ 1 \end{bmatrix}$

Where (x',y')are the coordinates in the rotated frame.

**5. Invert Color**

**Purpose**: Inverts the color of each frame. Color inversion is useful for testing models against unusual lighting conditions and color variability, ensuring robustness against diverse environmental conditions.

**Implementation**: This is achieved by subtracting each pixel value from 255 (i.e., 255 - frame), effectively reversing the color spectrum.

The color inversion for a pixel value p in frame F can be represented as: F inverted(i, j) = 255 - F(i, j)

**6. Downsample**

**Purpose**: Reduces the frame rate of the video by a specified factor. Downsampling is used to simulate lower frame rate videos and to test the efficiency and effectiveness of models under reduced temporal resolution conditions.

**Implementation**: The function selectively writes frames to the output video based on the downsample factor; for instance, writing every second frame for a factor of 2.

If the original frame rate is R and the downsample factor is $\beta$, the new frame rate R' is: $R' = \frac{R}{\beta}$

And frames are selected according to: $F'(t) = F(\beta \times t)$

Where t is the frame index in the downsampled video.

These augmentations introduce variability into the dataset, which is essential for training robust machine learning models, particularly in object detection and video analysis tasks. By applying these transformations, models can be trained to perform well across a variety of real-world conditions, ensuring they are not only accurate but also versatile in handling different scenarios and challenges.

## 3.3 Proposed Methodology

This study uses videos to show how to recognize and classify human behavior using a robust deep learning model. Videos, on the other hand, are collections of images put in a certain order to create motion. For video categorization, a variety of techniques might be used. This study suggested using VideoMAE along with YOLOv7 and video augmentation. We provide a complete explanation of the essential elements included in our suggested models in the parts that follow, along with a detailed display of their architectural layouts.

## 3.4 VideoMAE [38]

The intrinsic properties of video data, such as temporal redundancy—where successive frames often contain similar information—and temporal correlation—where events unfold over time—present unique opportunities and challenges for machine learning models. The Video Masked Autoencoder (VideoMAE) framework leverages these properties, utilizing a strategy adapted from successes in natural language processing (NLP) and still image analysis, where masked autoencoders have demonstrated significant , see fig 7.

### VideoMAE Architecture

#### Masking Strategy

– **Tube Masking**: Central to VideoMAE is its novel tube masking strategy, which applies a consistent mask across the temporal dimension of video clips. This approach ensures that if a segment is masked in one frame, the corresponding segments in adjacent frames are also masked. This strategy significantly increases the reconstruction challenge, preventing the model from merely interpolating spatially adjacent frames to fill in gaps, thus encouraging the learning of robust spatiotemporal features.

#### Encoder-Decoder Model

– **Asymmetric Architecture**: VideoMAE adopts an asymmetric encoder-decoder architecture, common in masked autoencoder designs but tailored for video. The heavy encoder processes the unmasked frames, transforming them into a rich, compressed latent space. Conversely, the lightweight decoder aims to reconstruct the original full video from this compressed representation.

– **Vision Transformer (ViT)**: Unlike many conventional video processing models that rely on 3D convolutions, VideoMAE utilizes a plain ViT, treating video frames as sequences of flattened 2D patches (tokens). This adaptation allows the model to leverage the powerful self-attention mechanisms of transformers, facilitating a global understanding of video content across both spatial and temporal dimensions.

### Methodology

#### Self-Supervised Pre-training

– **Training Objective**: The model is trained in a self-supervised manner by predicting the video parts that are masked out, using only the visible segments. The objective is to minimize the difference between the reconstructed video and the original unmasked video, typically measured by a reconstruction error such as Mean Squared Error (MSE).

– **High Masking Ratio**: VideoMAE employs an unusually high masking ratio (up to 95%), a strategy made viable by the redundancy in video data. This extensive masking forces the model to infer significant portions of the video, enhancing its ability to learn predictive and generative video features without relying on vast amounts of labeled data.

```
VideoMAEForVideoClassification(
  (videomae): VideoMAEModel(
    (embeddings): VideoMAEEmbeddings(
      (patch_embeddings): VideoMAEPatchEmbeddings(
        (projection): Conv3d(3, 768, kernel_size=(2, 16, 16), stride=(2, 16, 16))
      )
    )
    (encoder): VideoMAEEncoder(
      (layer): ModuleList(
        (0-11): 12 x VideoMAELayer(
          (attention): VideoMAEAttention(
            (attention): VideoMAESelfAttention(
              (query): Linear(in_features=768, out_features=768, bias=False)
              (key): Linear(in_features=768, out_features=768, bias=False)
              (value): Linear(in_features=768, out_features=768, bias=False)
              (dropout): Dropout(p=0.0, inplace=False)
            )
            (output): VideoMAESelfOutput(
              (dense): Linear(in_features=768, out_features=768, bias=True)
              (dropout): Dropout(p=0.0, inplace=False)
            )
          )
          (intermediate): VideoMAEIntermediate(
            (dense): Linear(in_features=768, out_features=3072, bias=True)
            (intermediate_act_fn): GELUActivation()
          )
          (output): VideoMAEOutput(
            (dense): Linear(in_features=3072, out_features=768, bias=True)
            (dropout): Dropout(p=0.0, inplace=False)
          )
          (layernorm_before): LayerNorm((768,), eps=1e-12, elementwise_affine=True)
          (layernorm_after): LayerNorm((768,), eps=1e-12, elementwise_affine=True)
        )
      )
    )
    (layernorm): LayerNorm((768,), eps=1e-12, elementwise_affine=True)
  )
  (classifier): Linear(in_features=768, out_features=3, bias=True)
)
```

**Fig.7.** VideoMAE Classifier Architecture

# 4 EXPERIMENTAL RESULT

The **High-Performance Computing (HPC) - SURYA** was used to train the model. The HPC infrastructure features 16 compute nodes, we utilized one HPC node equipped with 40 CPU cores, 2 NVIDIA V100 GPUs with 24GB RAM each, and 376GB of system RAM.

## 4.1 PERFORMANCE PARAMETERS

In the discipline of deep learning, a classification report serves as a statistical gauge of performance. Its goal is to show off the classification model's accuracy, recall, F1 score, precision, and loss metrics. For multiclass classification sparse-categorical-cross-entropy loss function is used as a loss function.

*Accuracy*: The accuracy of a model is determined by dividing the total number of predictions by the number of accurate predictions it made. The formula to calculate accuracy is as follows:

$$Accuracy = \frac{\sum_{i=1}^{N} I(y_i - \hat{y}_i)}{N}$$

Where: N is the total number of data, $y_i$ is true value of $i^{th}$ data, $\hat{y}_i$ is predicted value of $i^{th}$ data.

*Recall*: It is the ratio of accurately recognized instances of a given class to the total number of actual instances of that class for that specific class.

$$Recall_{class_i} = \frac{TP_{class_i}}{TP_{class_i} + FN_{class_i}}$$

*Precision*: It is defined as the ratio of successfully recognized instances of a given class to all instances projected to belong to that class.

$$Precision_{class_i} = \frac{TP}{TP_{class_i} + FP_{class_i}}$$

*F1 score*: The F1 Score is a balance between precision and recall, calculated as the harmonic mean of the two. When there is an uneven distribution of classes, it is very useful.

$$F1\ Score = 2 \times \frac{Precision_{class_i} \times Recall_{class_i}}{Precision_{class_i} + Recall_{class_i}}$$

*Loss*: The sparse-categorical-crossentropyloss function is best-suited loss function for multiclass classification. The negative log-likelihood of the expected probability of the true class is the loss for a single observation. The average of the various losses is the overall loss over the whole dataset (or over a batch during training):

$$Sparse\ Categorical\ Crossentropy = -\frac{1}{N}\sum_{i=1}^{N}\log(p_{i,y_i})$$

## 4.2 Ablation Studies

Our work involved an ablation methodology to evaluate the influence of various hyperparameters on the performance of the model. Through the systematic manipulation of batch size, split ratio, and learning rate, we successfully assessed the separate impacts of these factors on the overall efficacy of the model. The present study facilitated the identification of ideal configurations, therefore enhancing the performance of the model and guaranteeing the attainment of more robust and precise outcomes. This work provided vital insights that were essential for optimising the design of the model and directing its subsequent development.

In training machine learning models, especially those from the Hugging Face Transformers library, the max steps and epochs parameters both control the duration of the training process, but they do so in different ways.

**Epochs**: An epoch is a complete pass over the entire training dataset. The num train epochs parameter specifies how many times the training process should iterate over all the data in the dataset. Each epoch consists of several batches, depending on the size of the dataset and the batch size.

**Max Steps:** A training step is an iteration where a single batch of data is processed by the model. The **max steps** parameter specifies the total number of training steps (batches) to execute, overriding the **num train epochs** if the specified steps are reached first. This parameter provides a hard limit on the number of iterations, regardless of how many epochs have been completed.

**Usage Scenarios**

**When to Use Epochs:**

Epochs are used when you want the model to learn comprehensively from the entire dataset, often multiple times. It's beneficial when the focus is on ensuring that the model is exposed to all variations in the data across multiple passes.
It's particularly useful for smaller datasets where each pass through the data is relatively quick and essential for thorough learning.

**When to Use Max Steps:**

**Max steps** is ideal for very large datasets or when using streaming data where the dataset size might be infinite or undefined.
It's also useful in scenarios where training time is limited, and you need precise control over the training duration, independent of dataset size.
Helpful in experiments or when tuning hyperparameters to quickly assess the impact of changes without committing to full epochs of training.

**Interactions and Overrides**

If both **max steps** and **num _train epochs** are specified, **max steps** acts as an upper bound. This means the training will stop if the number of **max _steps** is reached, even if the model has not completed the number of epochs specified by **num train epochs.**

This overriding feature is particularly useful in avoiding excessively long training times especially in environments where computational resources are billed by usage or are time-limited.

**Practical Implications**

Resource Efficiency: Using max steps can be more resource-efficient in environments where you pay per compute time, as it prevents potentially unnecessary computations after sufficient learning has occurred.

In our case we are training the model for 100 epochs. But max step is running based on the formula *max steps* = (*train dataset.num videos//batch size*) ∗ *num epochs* ,where batch size = 4 and max step is overridding the epoch.

Table 6 presents the relationship between dataset size, maximum training steps, and epochs for different versions of the SSBD dataset. Although we set training to 100 epochs, the actual progression is governed by the formula of max step calculation. As the dataset expands—particularly in the trim SSBD yolov7 Aug version—the number of videos increases dramatically, which in turn raises the maximum steps and extends training beyond the nominal epoch count. This highlights how dataset scale and preprocessing choices directly influence the effective training cycle, ensuring the model is exposed to more diverse samples but also altering the dynamics of learning.

| Dataset | No. of videos for training | max steps calculated | Epochs |
|---|---|---|---|
| SSBD | 52 | 1300 steps | 171 |
| trim SSBD | 86 | 2100 steps | 190 |
| trim SSBD yolov7 | 86 | 2100 steps | 190 |
| trim SSBD yolov7 Aug | 606 | 15100 steps | 198 |

**Table 6.** Max steps vs Epochs

Table 7 compares the training times of the VideoMAE model across different versions of the SSBD dataset. Training on the original SSBD is the fastest (5.2 h), while preprocessing with trimming and Yolov7 slightly increases training time (7.5–8.9 h). The longest duration (51.1 h) is observed with Yolov7 plus augmentation, reflecting the additional computational cost of handling a significantly expanded dataset. This shows a clear trade-off between dataset enrichment and training efficiency.

**Table 7.** Training Time of VideoMAE model

| DataSet | Training Time |
|---|---|
| original SSBD | 5.2h |
| trim SSBD | 8.9 h |
| trim SSBD yolo7 | 7.5 h |
| trim SSBD yolo7 augmentation | 51.1 h |

### 4.2.1 For variation in Learning Rate on original SSBD data

Table 8 shows the effect of learning rate on the training time of the VideoMAE model using the SSBD dataset. A high learning rate (0.01) leads to the longest training time (37 h) due to unstable convergence, while reducing it to 0.001 lowers the time to 25.1 h. The shortest training times are observed at 0.0001 (7.86 h) and 0.00001 (7.81 h), indicating that very small learning rates offer no further gain. Thus, 0.0001 provides the best balance between efficiency and stability. Nevertheless, excessively low learning rates may risk underfitting, so both training speed and accuracy must be considered.

**Table 8.** Training Time of VideoMAE model with variation in learning rate

| Training Time | Learning Rate |
|---|---|
| 37 h | 0.01 |
| 25.1 h | 0.001 |
| 7.86 h | 0.0001 |
| 7.81 h | 0.00001 |

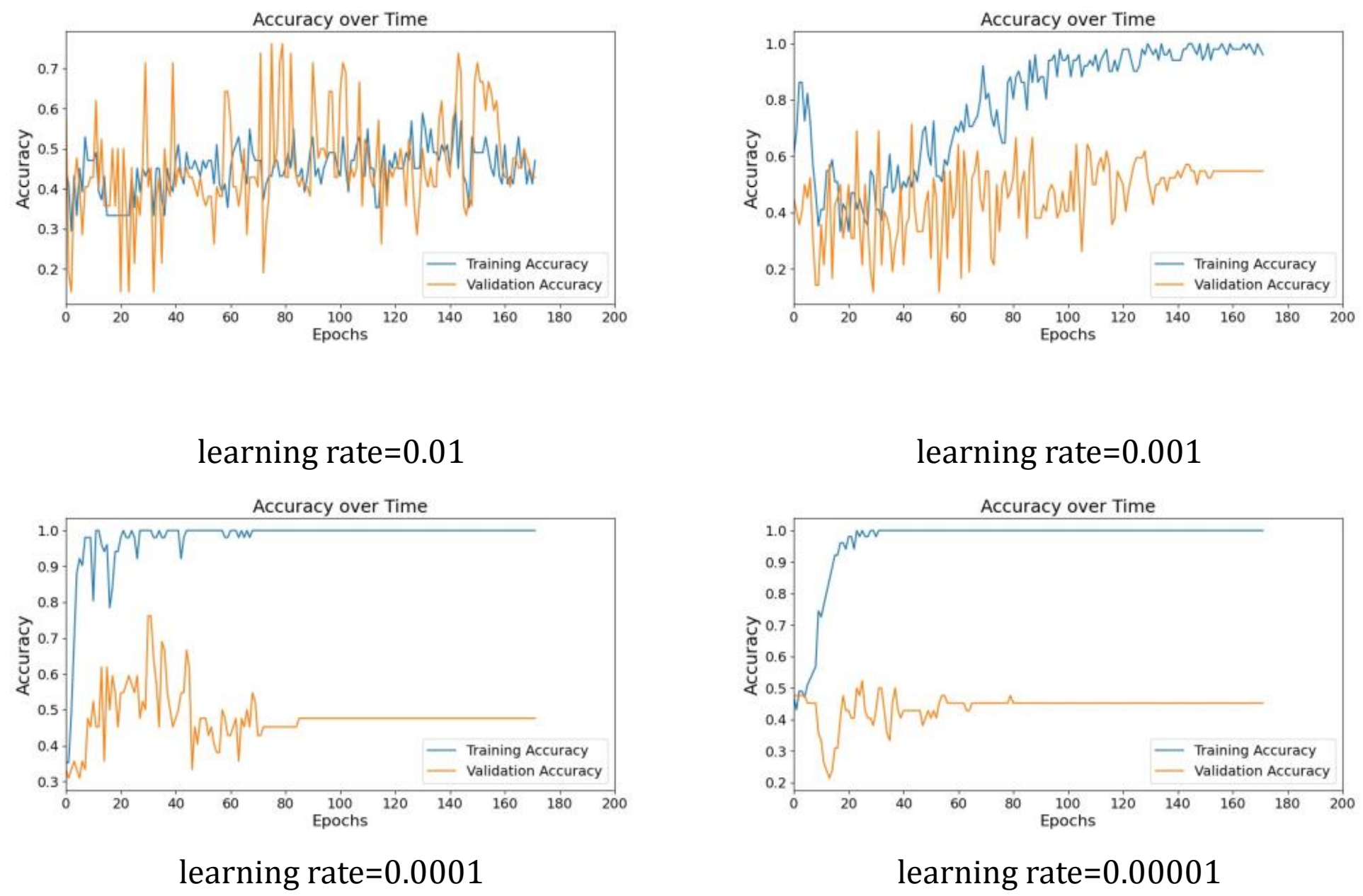

**Fig.8.** Comparison of Accuracy for various learning rate

Figure 8 presents a comparison of model accuracy over 200 epochs for various learning rates. With a learning rate of 0.01, both training and validation accuracies exhibit significant fluctuations, suggesting instability as the model overshoots optimal solutions. At a learning rate of 0.001, the training accuracy steadily improves and reaches high levels, while validation accuracy also increases but stabilizes lower, indicating some overfitting. At a learning rate of 0.0001, training accuracy quickly reaches high levels and remains stable, with validation accuracy gradually increasing, showing better generalization and less overfitting. However, with a learning rate of 0.00001, the training accuracy improves very slowly and validation accuracy remains low, indicating underfitting due to the excessively small learning rate. This analysis highlights the critical role of learning rate selection in balancing convergence speed and model performance.

Figure 9 compares the model's loss over 200 epochs for various learning rates, highlighting both training and validation losses. At a learning rate of 0.01, both losses initially decrease but fluctuate, indicating instability possibly due to excessive updates. With a learning rate of 0.001, the training loss steadily decreases, indicating effective learning, while validation loss fluctuates more, suggesting potential overfitting. At a learning rate of 0.0001, the training loss quickly drops and remains low, whereas the validation loss starts high and decreases slowly, reflecting better generalization. However, with a learning rate of 0.00001, the training loss decreases very slowly, and validation loss remains high, indicating underfitting due to the overly small learning rate. This analysis underscores the importance of selecting an appropriate learning rate to ensure effective model training and generalization.

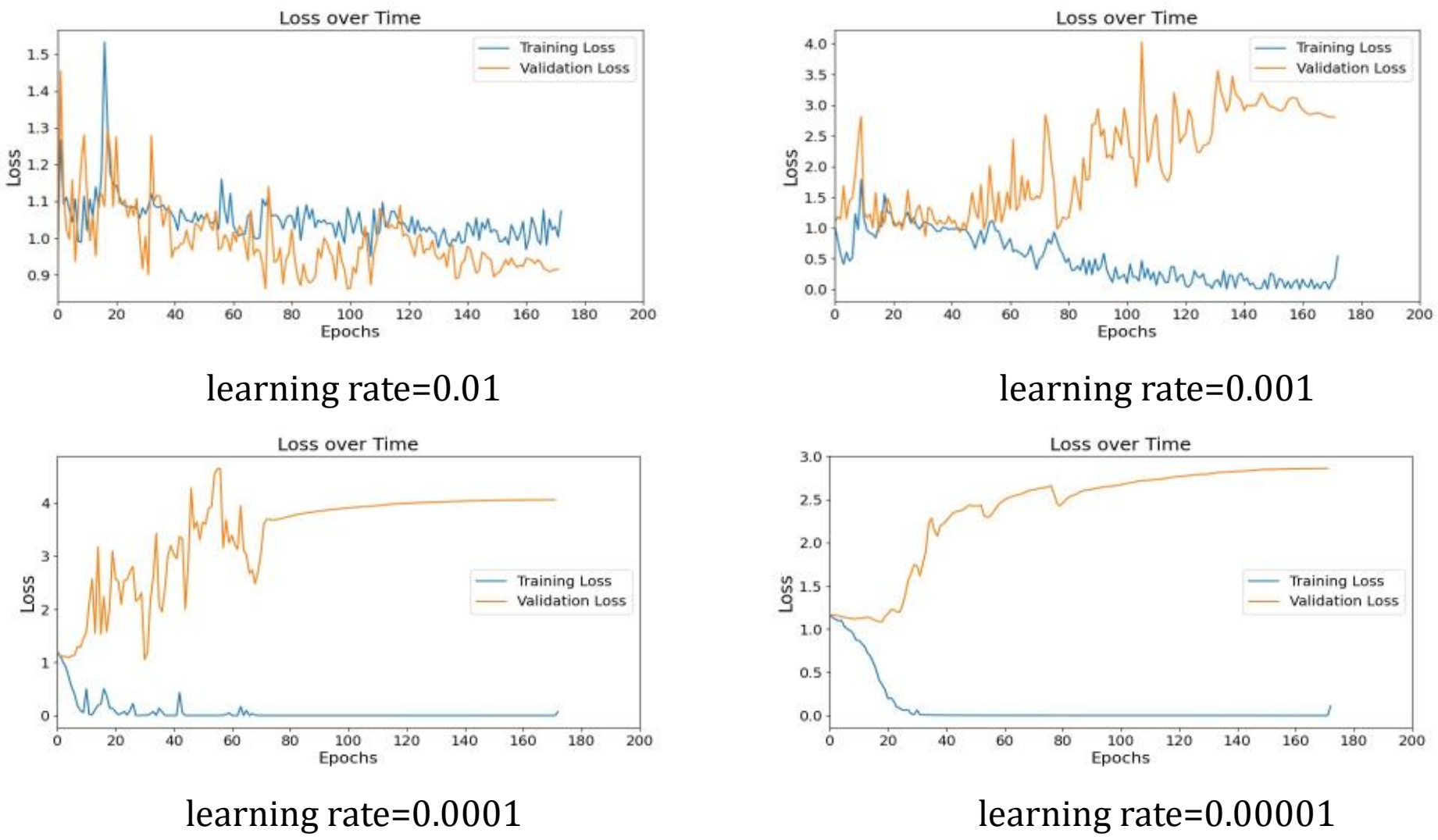

**Fig.9.** Comparison of Loss for various learning rate

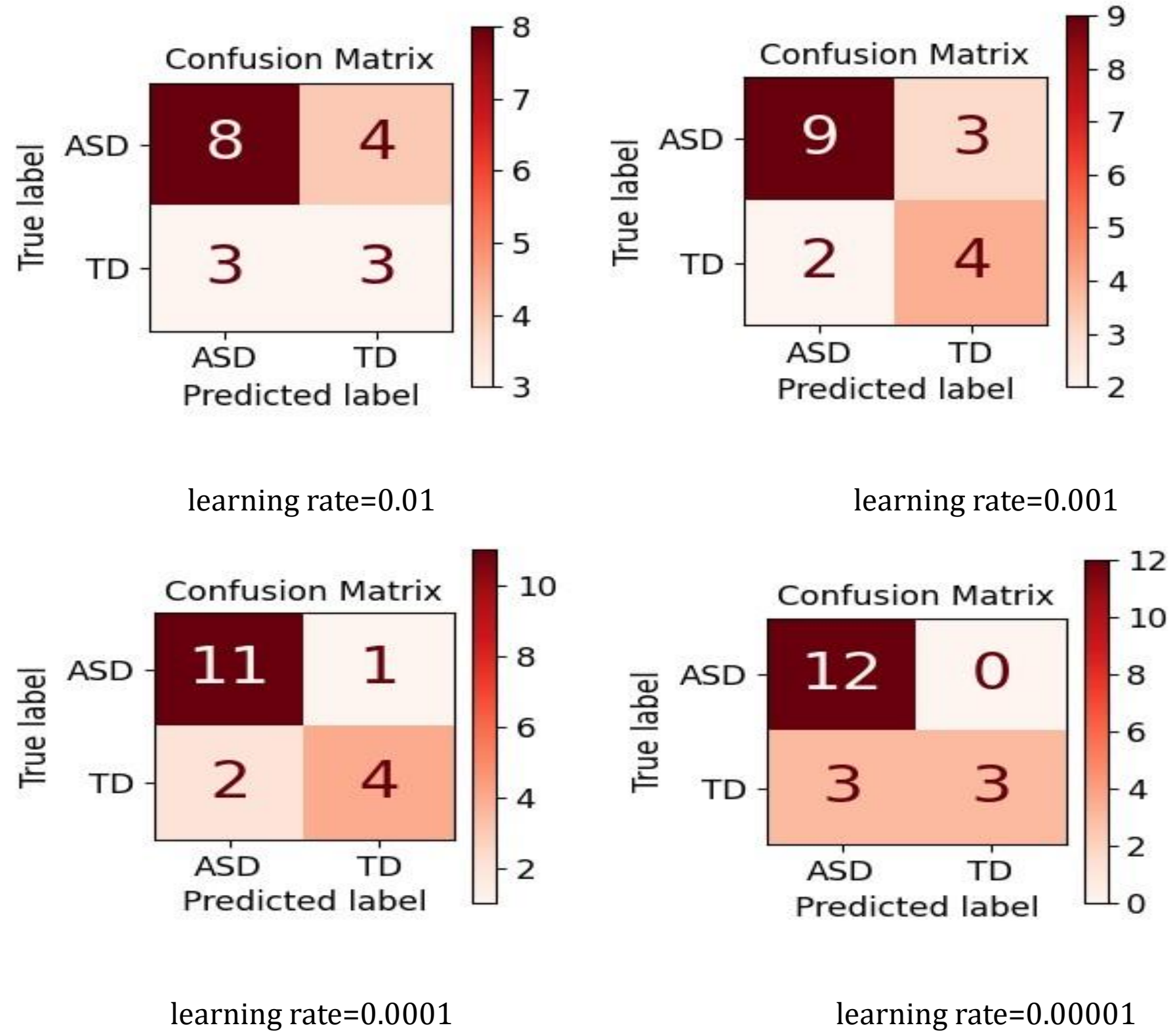

**Fig.10.** Comparison of Confusion Matrices

Figure 10 presents confusion matrices illustrating model performance across varying learning rates, demonstrating an overall improvement in accuracy as the learning rate increases. At a learning rate of 0.00001, the model shows high true positives for ASD but also some false positives, indicating good sensitivity but less specificity. As the learning rate rises to 0.0001, the model achieves a better balance by increasing true negatives and reducing false negatives, enhancing specificity. With a learning rate of 0.001, the model further balances false positives and false negatives, improving overall accuracy. At the highest learning rate of 0.01, while true positives and true negatives remain consistent, a slight increase in false classifications suggests potential overfitting or instability. This trend highlights the importance of optimizing learning rates to achieve the best model performance.

#### 4.2.2 For variation in Batch Size on Original SSBD data

Table 9 summarizes the training time of the VideoMAE model on the SSBD dataset under different batch size configurations. The results show a clear trend: increasing the batch size significantly reduces the overall training time. For example, training with a batch size of 2 requires 8.9 hours, while increasing the batch size to 16 reduces the time to 5.6 hours. This reduction occurs because larger batch sizes enable the model to process more samples in parallel, improving computational efficiency. However, while higher batch sizes accelerate training, they also demand greater GPU memory and may influence optimization stability and generalization performance. Therefore, an appropriate balance between training efficiency and model accuracy must be considered when selecting the batch size for VideoMAE on SSBD.

**Table 9.** Training Time of VideoMAE model with variation in learning rate

| Batch Size | Training Time for SSBD |
|---|---|
| 2 | 8.9 h |
| 4 | 7.7 h |
| 8 | 7.4 h |
| 16 | 5.6 h |

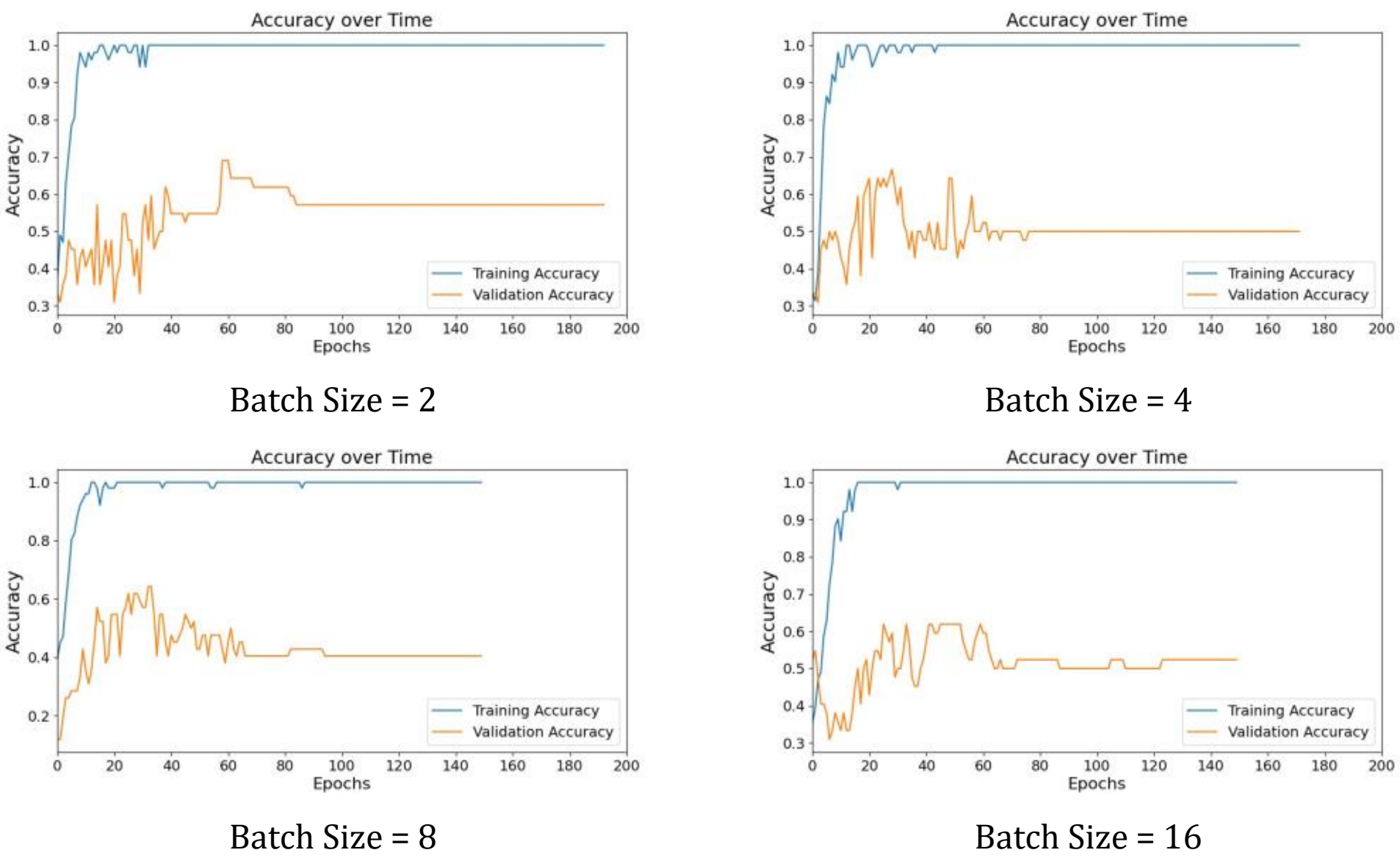

**Fig.11.** Comparison of Accuracy for various batch size

Figure 11 presents a comparison of training and validation accuracy across different batch sizes over 200 epochs. For a batch size of 2, training accuracy rapidly approaches near-perfect levels, while validation accuracy oscillates significantly, suggesting overfitting. At a batch size of 4, training accuracy remains high, but validation accuracy continues to fluctuate, indicating instability in generalization. With a batch size of 8, both training and validation accuracies exhibit smoother trends, yet validation accuracy stabilizes at a lower level, highlighting potential underfitting. At a batch size of 16, while training accuracy is consistently high, validation accuracy shows reduced fluctuations and stabilizes more effectively, suggesting a balance between model stability and generalization. This analysis underscores the importance of selecting an appropriate batch size to optimize model performance and generalization.

Figure 12 illustrates the comparison of training and validation loss over 200 epochs for various batch sizes. With a batch size of 2, training loss quickly decreases to near zero, while validation loss fluctuates and remains higher, indicating potential overfitting. For a batch size of 4, a similar pattern is observed, with training loss stabilizing at a low level and validation loss showing greater variability, suggesting instability in generalization. As the batch size increases to 8, training loss maintains a low value, but validation loss remains elevated and erratic, highlighting possible underfitting. At a batch size of 16, training loss consistently reaches minimal levels, while validation loss shows reduced fluctuations but stabilizes at a higher value than training loss, indicating an improved balance. This analysis emphasizes the need to carefully select batch sizes to achieve optimal model performance and generalization, balancing training efficiency and validation stability.

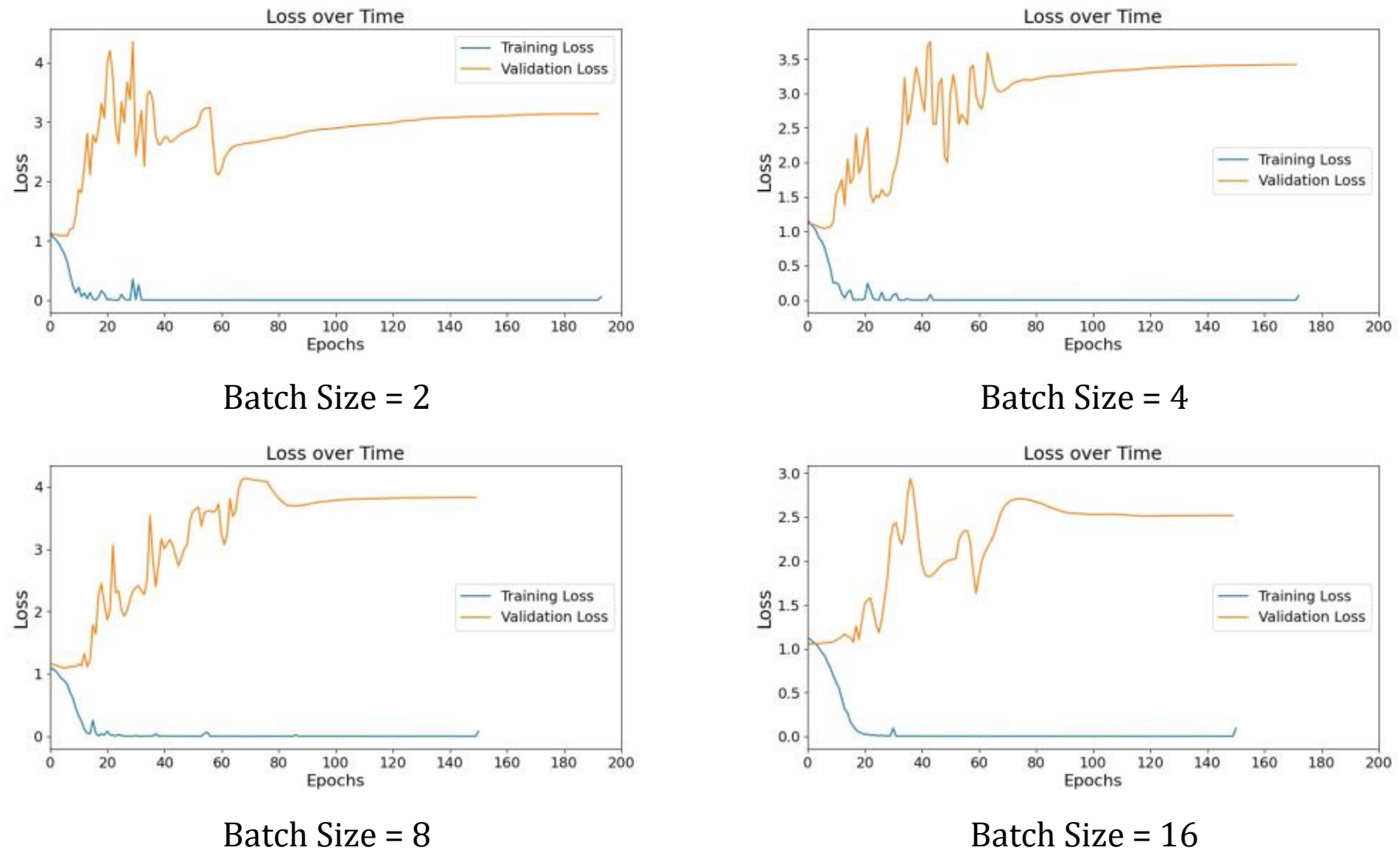

Batch Size = 2 Batch Size = 4

Batch Size = 8 Batch Size = 16

**Fig.12.** Comparison of Loss for various batch size

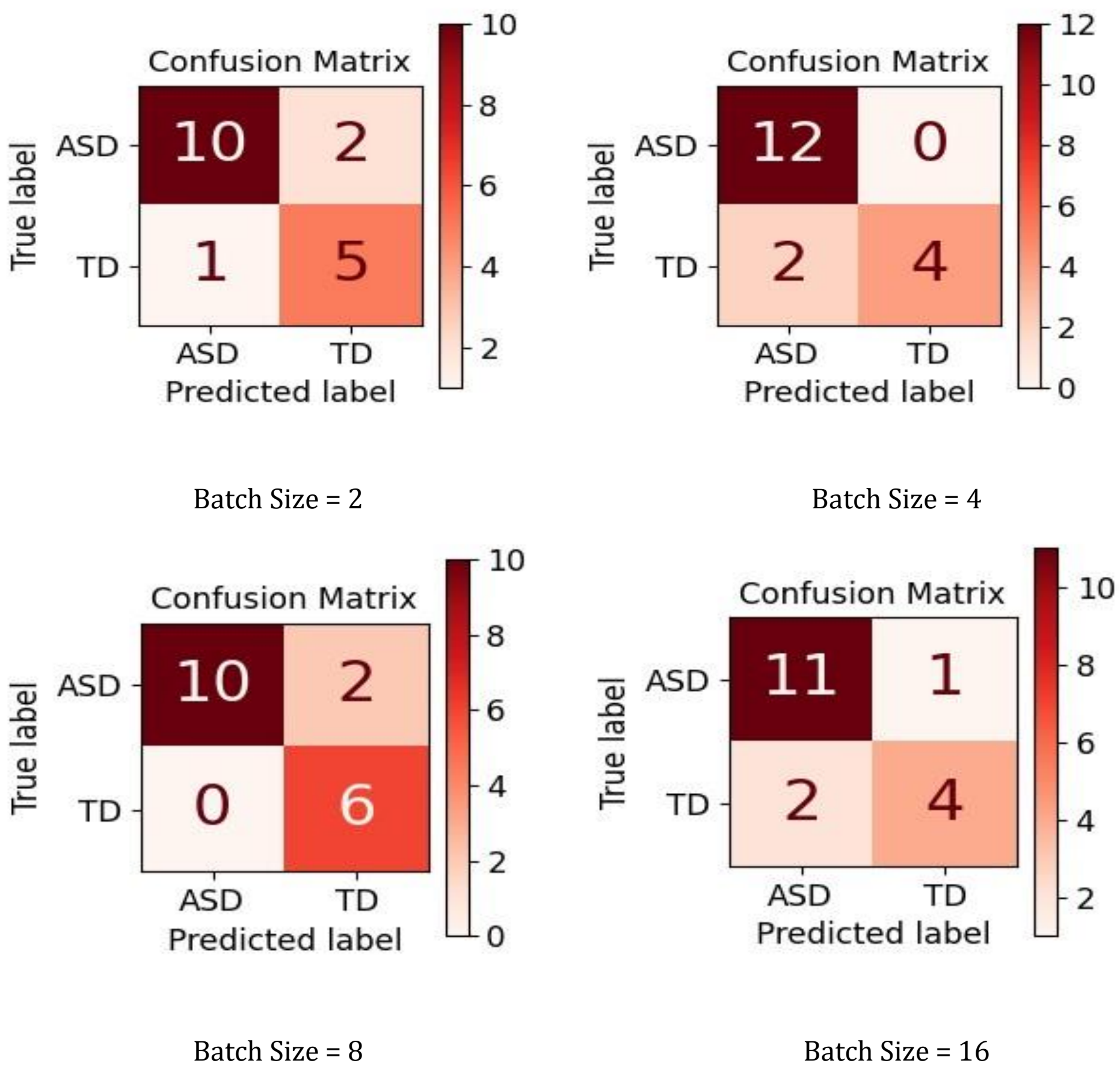

Batch Size = 2 Batch Size = 4

Batch Size = 8 Batch Size = 16

**Fig.13.** Comparison of Confusion Matrices for various batch size

Figure 13 displays confusion matrices comparing model performance across different batch sizes. For a batch size of 2, the model correctly identifies 10 ASD instances and 5 TD instances, but it misclassifies 2 ASD as TD and 1 TD as ASD, indicating slight misclassification. At a batch size of 4, the model improves, perfectly classifying all ASD instances and correctly identifying 4 TD instances, with 2 misclassifications of TD as ASD. With a batch size of 8, the model maintains consistent accuracy for ASD and improves TD classification, correctly identifying all 6 TD instances. At a batch size of 16, the model shows a strong performance, correctly classifying 11 ASD and 4 TD instances, with minor misclassification. This analysis demonstrates that adjusting batch size can significantly impact classification accuracy, with larger batch sizes generally improving performance and reducing misclassification.

### 4.2.3 For Variation in Split Ratio on Original SSBD data

**Table 10.** Training Time of VideoMAE model with variation in learning rate

| Split Ratio | Training Time for SSBD |
|---|---|
| 60:20:20 | 7.1 h |
| 70:15:15 | 7.7 h |
| 80:10:10 | 8.6 h |

The table 10 shows the training time of the VideoMAE model on the SSBD dataset with different split ratios. With a 60:20:20 split, the training time is 7.1 hours, providing a balanced data distribution. Increasing the training data to 70:15:15 raises the time to 7.7 hours, due to processing more data. At 80:10:10, the time further increases to 8.6 hours, reflecting the higher computational demand. This illustrates the trade-off between training time and data size, where more data enhances learning but requires more resources.

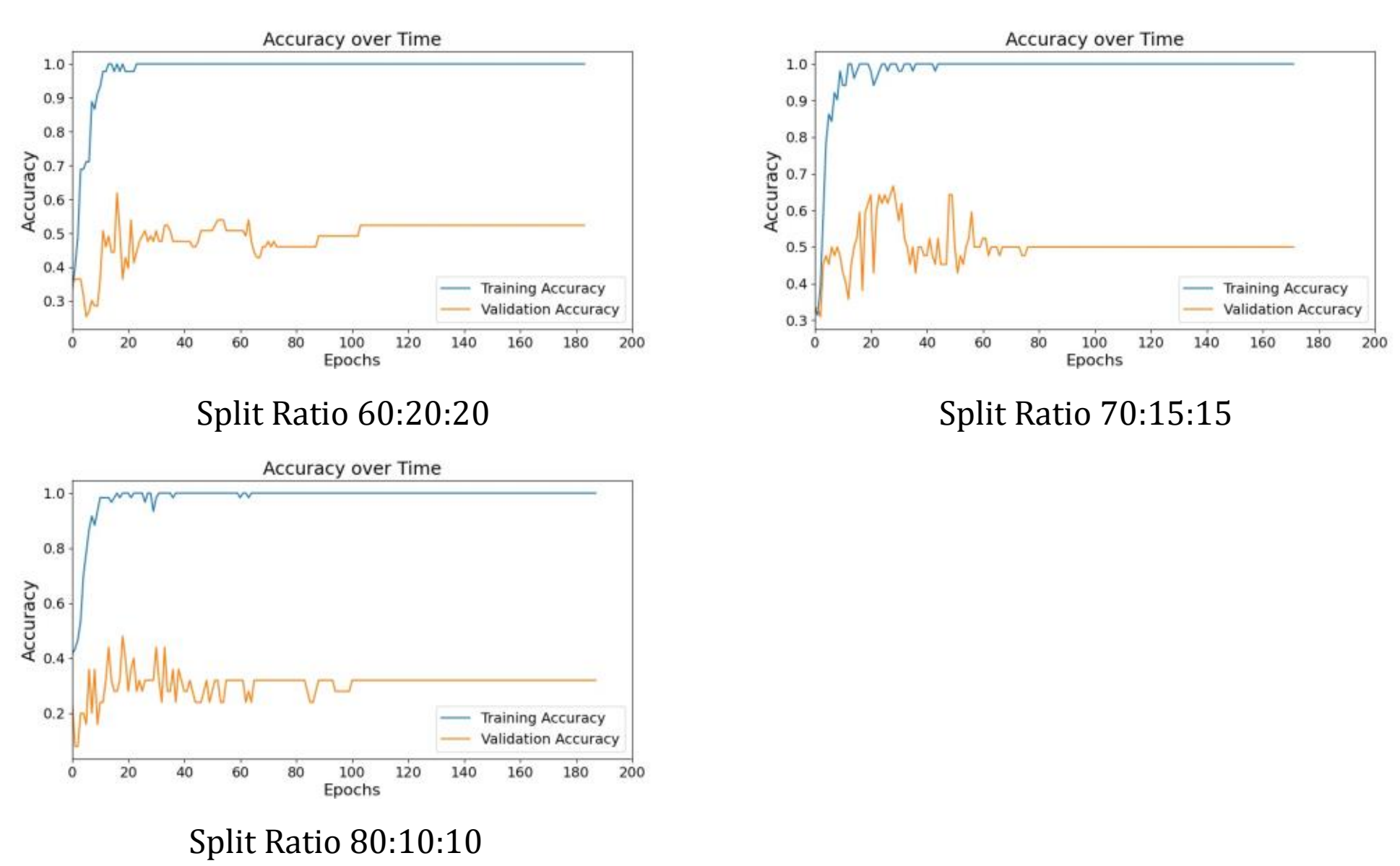

**Fig.14.** Comparison of Accuracy for various split ratio

The Fig 14 illustrate the VideoMAE model's training and validation accuracy across different data split ratios on the SSBD dataset over 200 epochs. In the 60:20:20 split, training accuracy quickly reaches nearperfect levels, around 0.98, while validation accuracy stabilizes around 0.7, indicating a potential overfitting issue where the model performs well on training data but less so on unseen data.

For the 70:15:15 split, training accuracy achieves 1.0 swiftly, indicating efficient learning from the larger training set. However, validation accuracy varies, stabilizing around 0.8. The model's ability to generalize slightly improves, reflecting a better balance between training and validation data.

In the 80:10:10 split, training accuracy also reaches 1.0, but validation accuracy remains at about 0.75. The increased training data enhances learning, yet the limited validation data may restrict the model's ability to generalize, highlighting the importance of a balanced data split.

Overall, these graphs demonstrate the trade-offs between training data size and model generalization. While larger training sets improve learning, adequate validation data is crucial to ensure robust performance on unseen data.

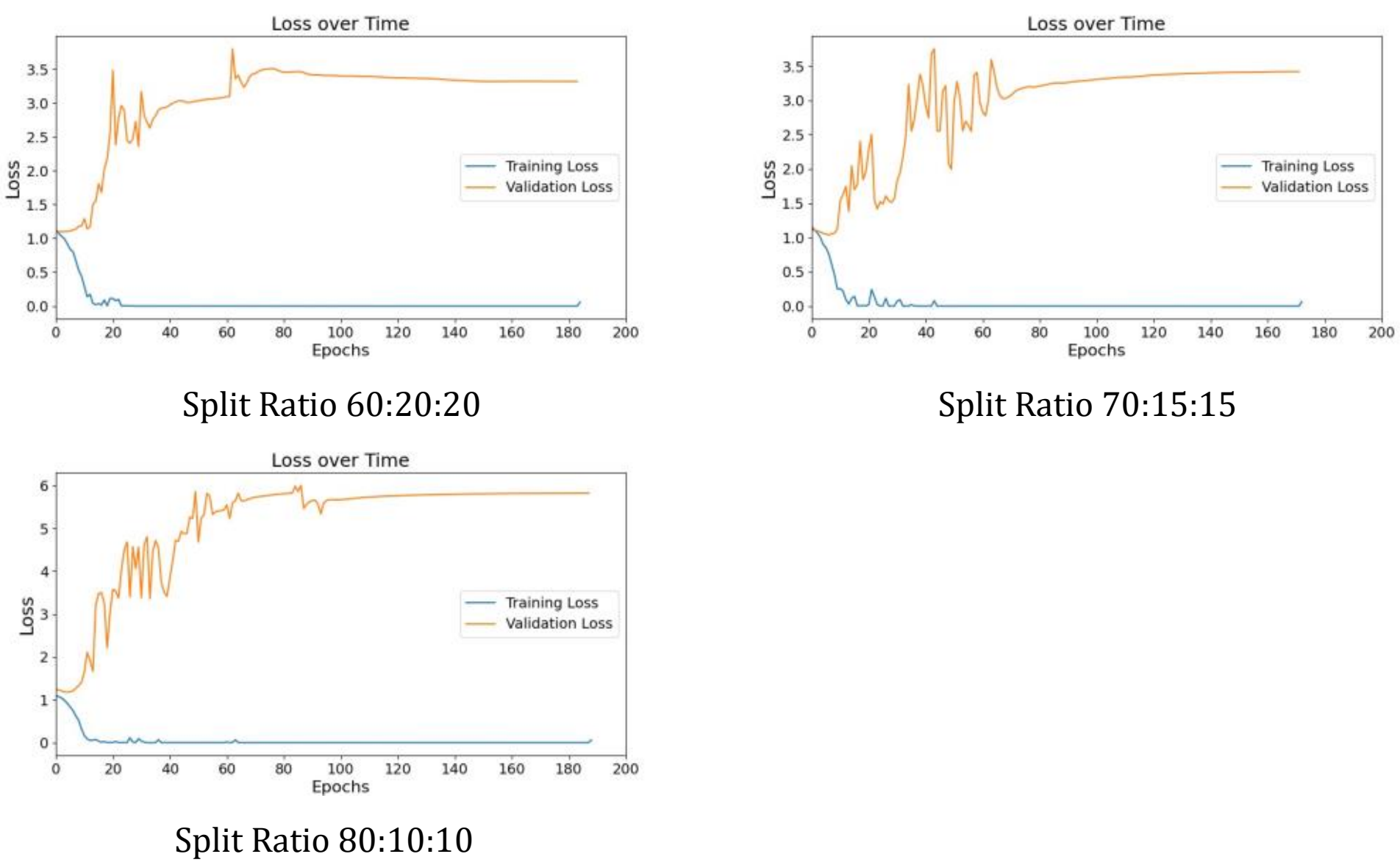

Split Ratio 60:20:20

Split Ratio 70:15:15

Split Ratio 80:10:10

**Fig.15.** Comparison of Loss for various split ratio

The Fig 15 depict the training and validation loss of the VideoMAE model with different split ratios on the SSBD dataset over 200 epochs. In the 60:20:20 split, training loss quickly drops to near zero, indicating effective learning. However, validation loss stabilizes around 3.5, suggesting overfitting, where the model performs well on training data but struggles to generalize to new data.

For the 70:15:15 split, training loss also decreases to nearly zero, but validation loss fluctuates before stabilizing around 3.0. This indicates slight improvement in generalization, although overfitting remains a concern due to the relatively large difference between training and validation loss.

In the 80:10:10 split, training loss again trends to zero, while validation loss stabilizes around 2.8. The increased training data enhances the model's learning, but limited validation data still leads to significant overfitting, as evidenced by the persistent gap between training and validation loss.

These graphs highlight the challenge of balancing training data size with validation data sufficiency. While more training data improves learning efficiency, adequate validation data is crucial to ensure the model's ability to generalize effectively to unseen data.

The confusion matrices in fig 16 illustrate the classification performance of the VideoMAE model on the SSBD dataset with different split ratios. In the 60:20:20 split, the model accurately classifies 12 ASD instances but misclassifies 3 as TD, indicating strong performance with some errors in distinguishing between classes. The model perfectly classifies all 8 TD instances.

In the 70:15:15 split, the model again correctly identifies 12 ASD instances and improves by not misclassifying any as TD. However, it predicts 4 TD instances correctly but misses 2, indicating improved specificity for ASD but some challenges with TD classification.

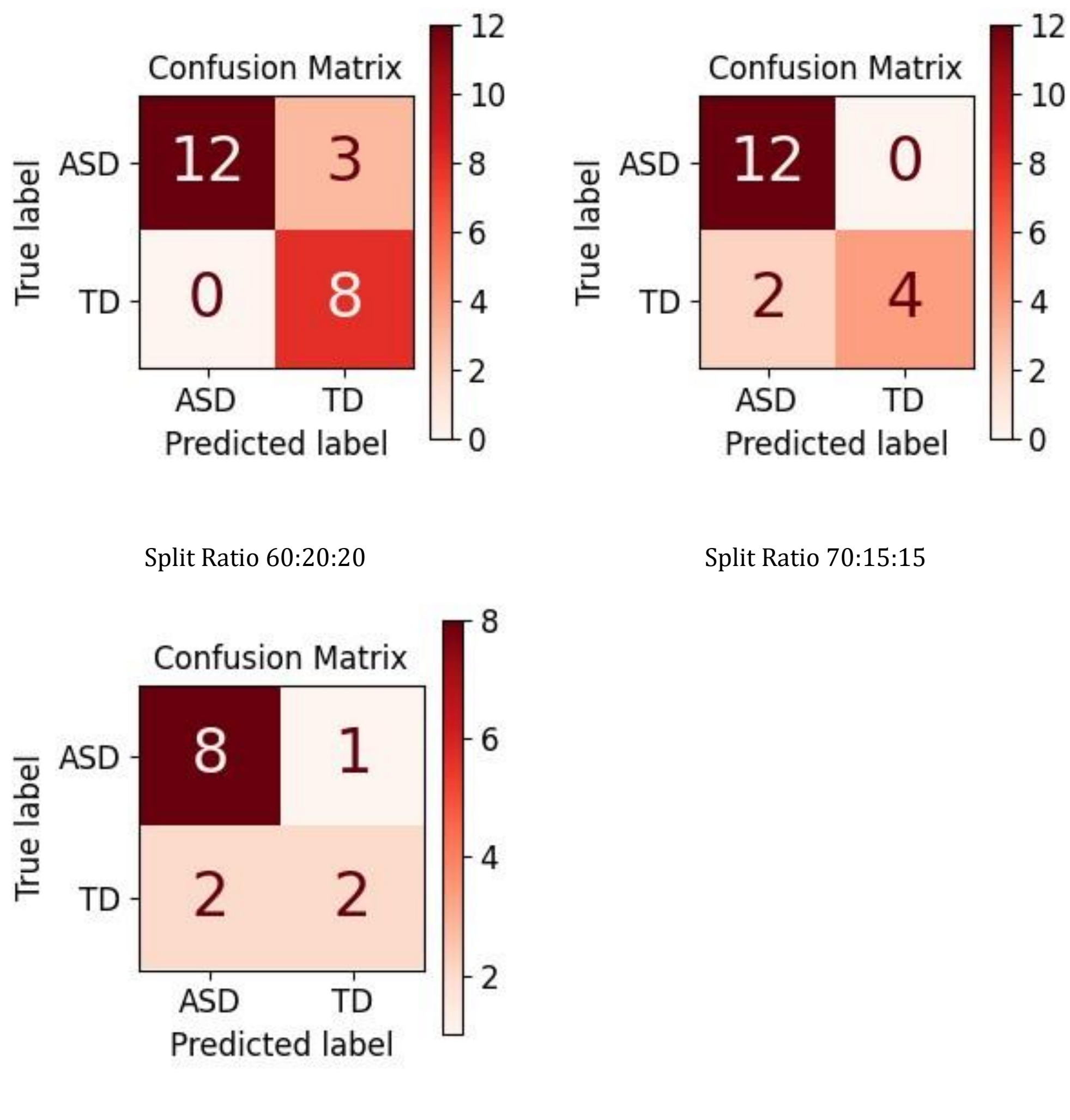

Split Ratio 80:10:10

**Fig.16.** Comparison of Confusion Matrices for various split ratio

For the 80:10:10 split, the model correctly classifies 8 ASD instances but makes more errors, misclassifying 6 as TD. It correctly identifies 2 TD instances but misclassifies 2 as ASD. This shows a decline in accuracy, likely due to the reduced validation set leading to overfitting.

Overall, the confusion matrices reveal that while increasing the training data can enhance ASD classification, maintaining a balanced split is crucial for accurate TD classification and overall generalization.

### 4.3 Results and Analysis

The SSBD dataset consists of total 75 videos, 25 video of each class. We randomly divide the dataset into training, testing and validation in the ratio of 70:15:15. Further, to validate the model at the time of training this 15% validation data is used. Training data has 17 videos of arm flapping, 17 videos of head banging, and 17 videos of spinning class. Testing data has 4 videos of arm flapping, 4 videos of head banging, and 4 videos of spinning class. Validation data has 4 videos of arm flapping, 4 videos of head banging, and 4 videos of spinning class.

The trim SSBD dataset consists of total 124 videos. We randomly divide the dataset into training, testing and validation in the ratio of 70:15:15. Further, to validate the model at the time of training this 15% validation data is used. Training data has 20 videos of arm flapping, 28 videos of head banging, and 37 videos of spinning class. Testing data has 5 videos of arm flapping, 7 videos of head banging, and 9 videos of spinning class. Validation data has 4 videos of arm flapping, 6 videos of head banging, and 8 videos of spinning class.

The trim SSBD yolo _augment dataset consists of total 868 videos. We randomly divide the dataset into training, testing and validation in the ratio of 70:15:15. Further, to validate the model at the time of training this 15% validation data is used. Training data has 142 videos of arm flapping, 200 videos of head banging, and 264 videos of spinning class. Testing data has 31 videos of arm flapping, 44 videos of head banging, and 57 videos of spinning class. Validation data has 30 videos of arm flapping, 43 videos of head banging, and 57 videos of spinning class.

**Table 11.** Class-wise Results of VideoMAE model on different SSBD variations

| Dataset | Class | Accuracy | Precision | Recall | F1 Score |
|---|---|---|---|---|---|
| Original SSBD | ArmFlapping | 1.00 | 1.00 | 1.00 | 1.00 |
| | HeadBanging | 0.50 | 1.00 | 0.50 | 0.66 |
| | Spinning | 1.00 | 0.66 | 1.00 | 0.80 |
| | TD | 0.67 | 1.00 | 0.66 | 0.80 |
| Trimmed SSBD | ArmFlapping | 0.40 | 0.40 | 0.40 | 0.40 |
| | HeadBanging | 0.28 | 0.50 | 0.28 | 0.36 |
| | Spinning | 1.00 | 0.75 | 1.00 | 0.85 |
| | TD | 0.83 | 1.00 | 0.83 | 0.90 |
| Trim SSBD yolo | ArmFlapping | 0.60 | 1.00 | 0.60 | 0.75 |
| | HeadBanging | 0.85 | 0.75 | 0.85 | 0.80 |
| | Spinning | 1.00 | 0.90 | 1.00 | 0.94 |
| | TD | 0.33 | 0.66 | 0.33 | 0.44 |
| SSBD yolo Augment | ArmFlapping | 0.93 | 1.00 | 0.93 | 0.96 |
| | HeadBanging | 0.97 | 0.97 | 0.97 | 0.97 |
| | Spinning | 1.00 | 0.96 | 1.00 | 0.98 |
| | TD | 0.97 | 1.00 | 0.97 | 0.98 |

The tables 11 illustrate the impact of different preprocessing techniques and datasets on the performance of the VideoMAE model, highlighting significant insights.

In the original SSBD dataset, ”ArmFlapping” achieves perfect classification, but ”HeadBanging” struggles with recall at 0.5, suggesting difficulty in correctly identifying all instances of this class. This indicates potential areas for model improvement or the need for more diverse training data.

The trimmed SSBD dataset results show a notable decline in performance for ”ArmFlapping” and ”HeadBanging,” while ”Spinning” remains stable. This suggests that trimming may remove essential contextual information, affecting model accuracy, especially for complex movements.

Introducing YOLOv7 with trim SSBD enhances ”HeadBanging” and ”Spinning” performance but negatively impacts ”TD,” which suggests that while object detection aids some actions, it may require further tuning for optimal results across all classes.

The integration of augmentation with YOLOv7 in the final table yields the highest accuracy and F1 scores across all classes, emphasizing the importance of diverse data augmentation. This approach not only improves the model's ability to generalize but also stabilizes performance, particularly in challenging classes like ”HeadBanging” and ”TD.” This insight suggests that combining augmentation and advanced detection techniques can significantly enhance model robustness and accuracy.

| Dataset | Class | Precision | Recall | F1-Score | Accuracy | Time |
|---|---|---|---|---|---|---|
| SSBD plus | ArmFlapping | 0.83 | 0.83 | 0.83 | 0.83 | 7.7 hr |
| | HeadBanging | 1.0 | 0.5 | 0.67 | 0.50 | |
| | Spinning | 1.0 | 1.0 | 1.0 | 1.0 | |
| | TD | 0.85 | 1.0 | 0.8 | 0.92 | |
| SSBD plus trim | ArmFlapping | 0.71 | 0.71 | 0.71 | 0.71 | 8.17 hr |
| | HeadBanging | 1.0 | 1.0 | 1.0 | 1.0 | |
| | Spinning | 1.0 | 0.67 | 0.8 | 0.67 | |
| | TD | 0.75 | 1.0 | 0.85 | 1.0 | |
| SSBD plus yolo | ArmFlapping | 1.0 | 0.85 | 0.92 | 0.86 | 7.4 hr |
| | HeadBanging | 1.0 | 1.0 | 1.0 | 1.0 | |
| | Spinning | 1.0 | 1.0 | 1.0 | 1.0 | |
| | TD | 0.85 | 1.0 | 0.92 | 1.0 | |
| Aug SSBD plus | ArmFlapping | 1.0 | 0.97 | 0.98 | 0.98 | 64.9 hr |
| | HeadBanging | 0.94 | 1 | 0.96 | 1 | |
| | Spinning | 1.0 | 0.94 | 0.97 | 0.95 | |
| | TD | 0.95 | 1.0 | 0.97 | 1.0 | |

**Table 12.** Performance Metrics for SSBD plus and its variations for 4 class classification

The table 12 provides a comprehensive evaluation of the SSBD plus dataset and its variations, focusing on the classification performance across four distinct classes: ArmFlapping, HeadBanging, Spinning, and TD. This analysis encompasses key metrics such as precision, recall, F1-score, accuracy, and the time required for training, reflecting the influence of various preprocessing strategies.

For the original SSBD plus dataset, ArmFlapping shows balanced metrics with precision, recall, and F1score all at 0.83, indicating a stable yet improvable performance, possibly through enhanced feature extraction techniques. HeadBanging, however, reveals a significant gap between precision and recall, achieving perfect precision but only 0.5 recall. This disparity suggests that while the model accurately identifies positive instances, it fails to detect all occurrences, impacting overall reliability. In contrast, Spinning achieves perfect scores across all metrics, highlighting the model's ability to effectively capture the distinctive features of this class. The TD class benefits from high recall at 1.0, ensuring most true positive cases are identified, though precision at 0.85 indicates potential for slight improvement.

When examining the SSBD plus trim dataset, ArmFlapping experiences a drop to 0.71 across all metrics, underscoring the potential loss of essential features due to trimming, which may impede the model's generalization capabilities. Despite this, HeadBanging and TD maintain high performance, achieving perfect recall and accuracy, suggesting these classes are less affected by data reduction. Spinning shows a recall drop to 0.67, indicating some instances are missed, which could be mitigated through targeted data enhancement strategies.

The SSBD plus yolo dataset demonstrates significant performance improvements, particularly for ArmFlapping, with an F1-score of 0.92, illustrating the effectiveness of YOLO in capturing intricate patterns and enhancing feature extraction. Both HeadBanging and Spinning maintain perfect scores, reflecting YOLO's capability in ensuring high accuracy across varying movement patterns. Additionally, the reduction in training time to 7.4 hours highlights improved computational efficiency alongside these performance gains.

In the Aug SSBD plus dataset, ArmFlapping and TD achieve near-perfect metrics, with F1-scores of 0.98 and 0.97, respectively, showcasing the robustness of data augmentation in enhancing model performance and handling input variability. Although Spinning experiences a slight drop in recall to 0.94, it maintains a high F1-score of 0.97, indicating effective overall performance. However, the extensive augmentation process introduces a significant computational cost, with training duration extending to 64.9 hours. Despite this, the enhanced metrics across all classes underscore the critical role of augmentation, suggesting that the trade-off in training time is justified by the superior generalization and accuracy achieved.

These insights collectively reveal the nuanced impacts of data preprocessing and augmentation on model performance, offering a strategic roadmap for optimizing classification tasks in similar datasets. The findings highlight the importance of balancing data preprocessing strategies with computational resources to achieve the desired performance outcomes.

| **Dataset** | **Class** | **Precision** | **Recall** | **F1-Score** | **Accuracy** |
|---|---|---|---|---|---|
| SSBD plus | ASD | 1.0 | 0.90 | 0.95 | 0.91 |
| | TD | 0.85 | 1.0 | 0.92 | 1.00 |
| SSBD plus trim | ASD | 1.0 | 0.87 | 0.93 | 0.88 |
| | TD | 0.75 | 1.0 | 0.85 | 1.00 |
| SSBD plus yolo | ASD | 1.0 | 0.93 | 0.96 | 0.94 |
| | TD | 0.85 | 1.0 | 0.92 | 1.00 |
| Aug SSBD plus | ASD | 1.0 | 1 | 0.97 | 0.98 |
| | TD | 0.95 | 1.0 | 0.97 | 1.00 |

**Table 13.** Performance Metrics for SSBD plus and its variations for ASD-TD classification

The table 13 presents a comparison of classification performance across four variations of the SSBD plus dataset, focusing on the ASD (Autism Spectrum Disorder) and TD (Typically Developing) classes. Key performance metrics such as precision, recall, F1-score, and accuracy are used to evaluate each dataset's effectiveness.

For the SSBD plus dataset, the ASD class achieves perfect precision but with a recall of 0.90, suggesting that while false positives are minimized, some true ASD instances are missed. This results in an F1-score of 0.95 and an accuracy of 0.91, indicating strong but slightly improvable performance. The TD class has a precision of 0.85, with perfect recall ensuring all true instances are captured, resulting in an F1-score of 0.92 and flawless accuracy. This highlights the model's efficiency in identifying TD cases while maintaining high overall accuracy.

Moving to the SSBD plus trim dataset, the ASD class maintains perfect precision but experiences a reduction in recall to 0.87, leading to an F1-score of 0.93 and an accuracy of 0.88. This drop suggests that trimming may lead to the loss of important features, impacting recall. For the TD class, precision decreases to 0.75, reflecting a higher rate of false positives, yet recall remains perfect. This results in an F1-score of 0.85 and perfect accuracy, indicating effective detection of all TD cases despite precision challenges.

In the SSBD plus yolo dataset, improvements are seen in the ASD class with recall increasing to 0.93, while precision remains at 1.0. This enhances the F1-score to 0.96 and accuracy to 0.94, demonstrating YOLO's ability to better capture ASD features. The TD class maintains consistent performance with a precision of 0.85 and perfect recall, resulting in an F1-score of 0.92 and perfect accuracy, indicating stable and reliable classification.

The Aug SSBD plus dataset stands out with exceptional performance. The ASD class achieves perfect precision and recall, leading to an F1-score of 0.97 and accuracy of 0.98, reflecting the profound impact of data augmentation in improving model generalization and classification accuracy. Similarly, the TD class shows near-perfect metrics, with precision at 0.95 and perfect recall, achieving an F1-score of 0.97 and accuracy of 1.00. This underscores the effectiveness of augmentation in enhancing performance across both classes.

Overall, the analysis reveals the significant benefits of data augmentation, as seen in the Aug SSBD plus dataset, which consistently provides the most robust and accurate results. It highlights the importance of choosing appropriate preprocessing techniques to optimize classification outcomes, with data augmentation demonstrating the greatest potential for improving both precision and recall across ASD and TD classifications.

| Dataset | Class | Precision | Recall | F1-Score | Accuracy | Time |
|---|---|---|---|---|---|---|
| Modified SSBD | ArmFlapping | 0.0 | 0.0 | 0.0 | 0.0 | 7.7 hr |
| | HeadBanging | 0.67 | 0.95 | 0.79 | 0.95 | |
| | Spinning | 0.63 | 0.43 | 0.51 | 0.44 | |
| | TD | 1.0 | 0.16 | 0.28 | 0.17 | |
| Modified SSBD trim | ArmFlapping | 0.66 | 0.85 | 0.75 | 0.86 | 8.17 hr |
| | HeadBanging | 0.83 | 0.71 | 1.0 | 0.71 | |
| | Spinning | 0.9 | 0.9 | 0.9 | 0.90 | |
| | TD | 1.0 | 0.83 | 0.85 | 0.83 | |
| Modified SSBD yolo | ArmFlapping | 1.0 | 0.85 | 0.92 | 0.86 | 7.4 hr |
| | HeadBanging | 0.83 | 0.71 | 0.76 | 0.71 | |
| | Spinning | 0.81 | 0.9 | 0.85 | 0.90 | |
| | TD | 0.85 | 1.0 | 0.92 | 1.0 | |
| Modified SSBD Augment | ArmFlapping | 0.93 | 0.97 | 0.95 | 0.98 | 64.9 hr |
| | HeadBanging | 0.97 | 0.88 | 0.92 | 0.89 | |
| | Spinning | 0.96 | 0.92 | 0.94 | 0.92 | |
| | TD | 0.88 | 1.0 | 0.93 | 1.0 | |

**Table 14.** Performance Metrics for Modified SSBD and its variations for 4 class classification

The table 14 provides a detailed overview of the classification performance across four variations of the Modified SSBD dataset, focusing on four classes: ArmFlapping, HeadBanging, Spinning, and TD (Typically Developing). It uses metrics such as precision, recall, F1-score, accuracy, and training time to evaluate each dataset's efficiency.

In the Modified SSBD dataset, ArmFlapping shows zero performance across all metrics, indicating a complete inability to classify this class. HeadBanging achieves a precision of 0.67 and recall of 0.95, resulting in an F1-score of 0.79 and an accuracy of 0.95, highlighting good recall but room for precision improvement. Spinning has a precision of 0.63 and recall of 0.43, leading to an F1-score of 0.51 and an accuracy of 0.44, reflecting difficulties in capturing this class effectively. The TD class exhibits perfect precision but very low recall at 0.16, resulting in an F1-score of 0.28 and accuracy of 0.17, suggesting challenges in detecting true positives.

The Modified SSBD trim dataset shows improvement for ArmFlapping, with precision at 0.66, recall at 0.85, and an F1-score of 0.75, indicating better balance and accuracy of 0.86. HeadBanging achieves perfect F1-score at 1.0, though precision and recall could be more balanced. Spinning maintains high performance with all metrics at 0.9, demonstrating effective classification. The TD class has perfect precision and a recall of 0.83, resulting in an F1-score of 0.85 and accuracy of 0.83, indicating reliable classification with some room for recall improvement.

In the Modified SSBD yolo dataset, ArmFlapping achieves perfect precision and improved recall at 0.85, resulting in an F1-score of 0.92 and accuracy of 0.86, demonstrating significant enhancement. HeadBanging shows a balanced performance with precision at 0.83 and recall at 0.71, yielding an F1-score of 0.76. Spinning maintains high performance with an F1-score of 0.85, and the TD class achieves perfect recall, precision of 0.85, and an F1-score of 0.92, resulting in perfect accuracy.

The Modified SSBD Augment dataset excels across all classes, particularly ArmFlapping, with precision at 0.93, recall at 0.97, and an F1-score of 0.95, achieving accuracy of 0.98. HeadBanging and Spinning both show strong performance with F1-scores of 0.92 and 0.94, respectively, reflecting effective classification. The TD class maintains high performance with an F1-score of 0.93 and perfect accuracy. However, this improvement comes with a significant increase in training time to 64.9 hours, highlighting the computational cost of data augmentation.

Overall, the table demonstrates the varying impacts of dataset modifications on classification performance, with data augmentation in the Modified SSBD Augment dataset providing the most robust results across all classes. This underscores the value of data preprocessing techniques such as augmentation in enhancing model accuracy and generalization, despite increased computational demands.

The table 15 presents a comparative analysis of classification performance across four variations of the Modified SSBD dataset, focusing on two classes: ASD (Autism Spectrum Disorder) and TD (Typically

| Dataset | Class | Precision | Recall | F1-Score | Accuracy |
|---|---|---|---|---|---|
| Modified SSBD | ASD | 0.93 | 1.0 | 0.96 | 1.0 |
| | TD | 1.0 | 0.16 | 0.28 | 0.17 |
| Modified SSBD trim | ASD | 0.96 | 1.0 | 0.97 | 1.0 |
| | TD | 1.0 | 0.83 | 0.90 | 0.83 |
| Modified SSBD yolo | ASD | 1.0 | 0.95 | 0.97 | 0.96 |
| | TD | 0.85 | 1.0 | 0.92 | 1.00 |
| Modified SSBD _Augment | ASD | 1.0 | 0.96 | 0.98 | 0.97 |
| | TD | 0.88 | 1.0 | 0.93 | 1.00 |

**Table 15.** Performance Metrics for Modified SSBD and its variations for ASD TD classification

Developing). Key performance metrics such as precision, recall, F1-score, and accuracy are used to assess the efficacy of each dataset variation.

In the Modified SSBD dataset, the ASD class demonstrates strong performance with a precision of 0.93 and perfect recall of 1.0, resulting in an F1-score of 0.96 and full accuracy. This indicates a highly effective model for identifying ASD cases. However, the TD class shows a significant disparity with perfect precision but a low recall of 0.16, leading to an F1-score of 0.28 and accuracy of 0.17. This highlights challenges in detecting true positives within the TD class, suggesting a need for improved recall strategies.

The Modified SSBD trim dataset enhances performance for both classes. The ASD class achieves even higher precision at 0.96 and maintains perfect recall, resulting in an F1-score of 0.97 and full accuracy, indicating robust classification. The TD class sees substantial improvement, with perfect precision and recall rising to 0.83, leading to an F1-score of 0.90 and accuracy of 0.83. This reflects a more balanced and effective detection of TD instances, addressing previous recall challenges.

In the Modified SSBD yolo dataset, the ASD class achieves perfect precision and maintains high recall at 0.95, resulting in an F1-score of 0.97 and accuracy of 0.96. This demonstrates YOLO's capability to improve recall while maintaining perfect precision. The TD class achieves perfect recall and excellent precision at 0.85, with an F1-score of 0.92 and full accuracy, showcasing effective classification outcomes.

The Modified SSBD Augment dataset excels across both classes. The ASD class achieves perfect precision and high recall at 0.96, resulting in an F1-score of 0.98 and accuracy of 0.97, indicating outstanding classification performance. Similarly, the TD class demonstrates strong performance with precision at 0.88 and perfect recall, leading to an F1-score of 0.93 and full accuracy. This underscores the significant benefits of data augmentation in enhancing model performance, particularly in achieving high precision and recall.

Overall, the table illustrates the varying impacts of dataset modifications on classification performance, with data augmentation in the Modified SSBD Augment dataset yielding the most robust and accurate results across both classes. This highlights the importance of data preprocessing techniques, such as trimming and augmentation, in improving model accuracy and generalization, ultimately enhancing the classification outcomes for ASD and TD classes.

### 4.3.1 ASD and TD Result Comparision

**Table 16.** Class-wise Model Performance

| | SSBD | Aug SSBD | Modified SSBD | Aug Modified SSBD |
|---|---|---|---|---|
| ArmFlapping | 0.50 | 0.90 | 0.00 | 0.90 |
| HeadBanging | 0.50 | 1.0 | 0.95 | 1.00 |
| Spinning | 1.00 | 0.91 | 0.44 | 0.91 |
| TD | 0.67 | 0.97 | 0.17 | 0.97 |

The table 16 details the model's performance in identifying specific behaviors across different versions of the SSBD dataset, including augmented and modified variations. For "Arm Flapping," the model's accuracy significantly improves from 0.50 in the original SSBD to 0.90 in both the Augmented and Augmented Modified datasets, while it drops to 0.00 in the Modified SSBD. This underscores the positive impact of data augmentation on enhancing detection capabilities.

In "Head Banging," the model shows marked improvement, achieving perfect accuracy (1.00) in the Augmented and Augmented Modified datasets, up from 0.50 in the original SSBD and 0.95 in the Modified SSBD. This suggests that both augmentation and modifications contribute positively to classification accuracy.

For "Spinning," the model initially detects the behavior perfectly (1.00) in the original SSBD, but performance drops to 0.44 in the Modified SSBD, indicating possible issues with modifications. However, augmentation restores performance to 0.91, illustrating its role in mitigating negative effects of modifications.

Typically Developing (TD) behaviors see an accuracy increase from 0.67 in the SSBD to 0.97 in both the Augmented datasets, while dropping to 0.17 in the Modified SSBD. This pattern highlights challenges with modifications alone and the subsequent improvement with augmentation.

Overall, the data demonstrates that data augmentation significantly boosts the model's detection capabilities across behaviors, particularly when modifications negatively impact performance, emphasizing the importance of strategic preprocessing.

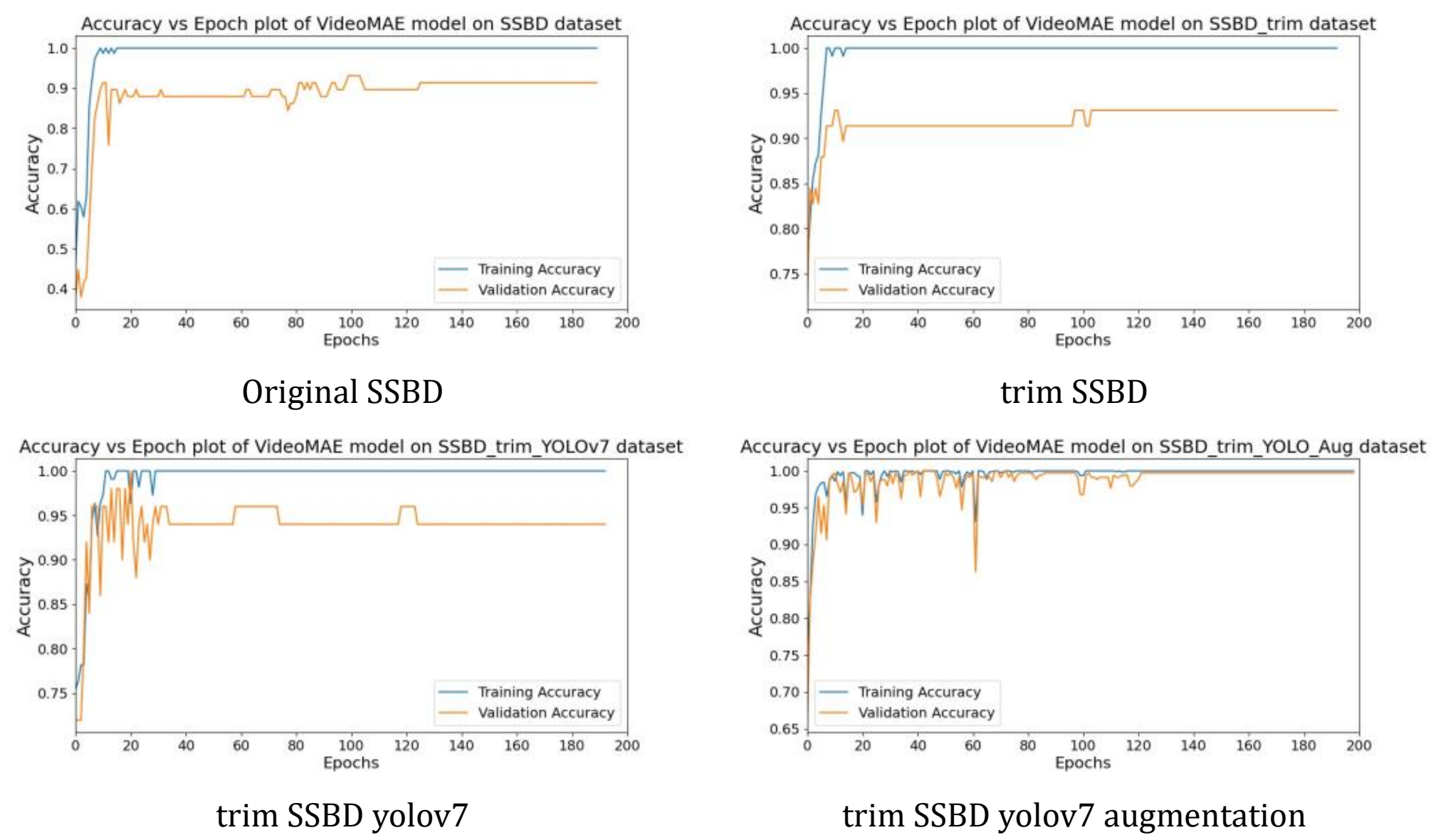

Original SSBD trim SSBD

trim SSBD yolov7 trim SSBD yolov7 augmentation

**Fig.17.** Comparison of Accuracy

The Fig 17 illustrate the training and validation accuracy of the VideoMAE model across various preprocessing techniques on the SSBD dataset, offering insights into the model's performance over 200 epochs.

In the Original SSBD graph, the model quickly reaches high accuracy, with both training and validation stabilizing around 0.95, indicating an effective baseline performance.

For the trim SSBD dataset, the model achieves slightly improved accuracy, with both training and validation curves quickly converging to around 0.97. This suggests that trimming enhances the model's ability to generalize by focusing on relevant features.

The trim SSBD yolo7 dataset shows a similar trend, with training and validation accuracy reaching approximately 0.98. The consistency between training and validation curves indicates a reduction in overfitting, likely due to the YOLO-based trimming emphasizing key features.

In the trim SSBD yolo7 augmentation graph, the model achieves the highest accuracy, with both training and validation curves reaching near-perfect accuracy close to 1.0. The use of augmentation appears to significantly boost the model's performance by providing a more diverse set of training examples, leading to enhanced generalization and robustness.

Overall, these graphs demonstrate that preprocessing techniques, particularly YOLO-based trimming and augmentation, substantially improve the VideoMAE model's accuracy on the SSBD dataset, highlighting the importance of strategic data preparation in optimizing model performance.

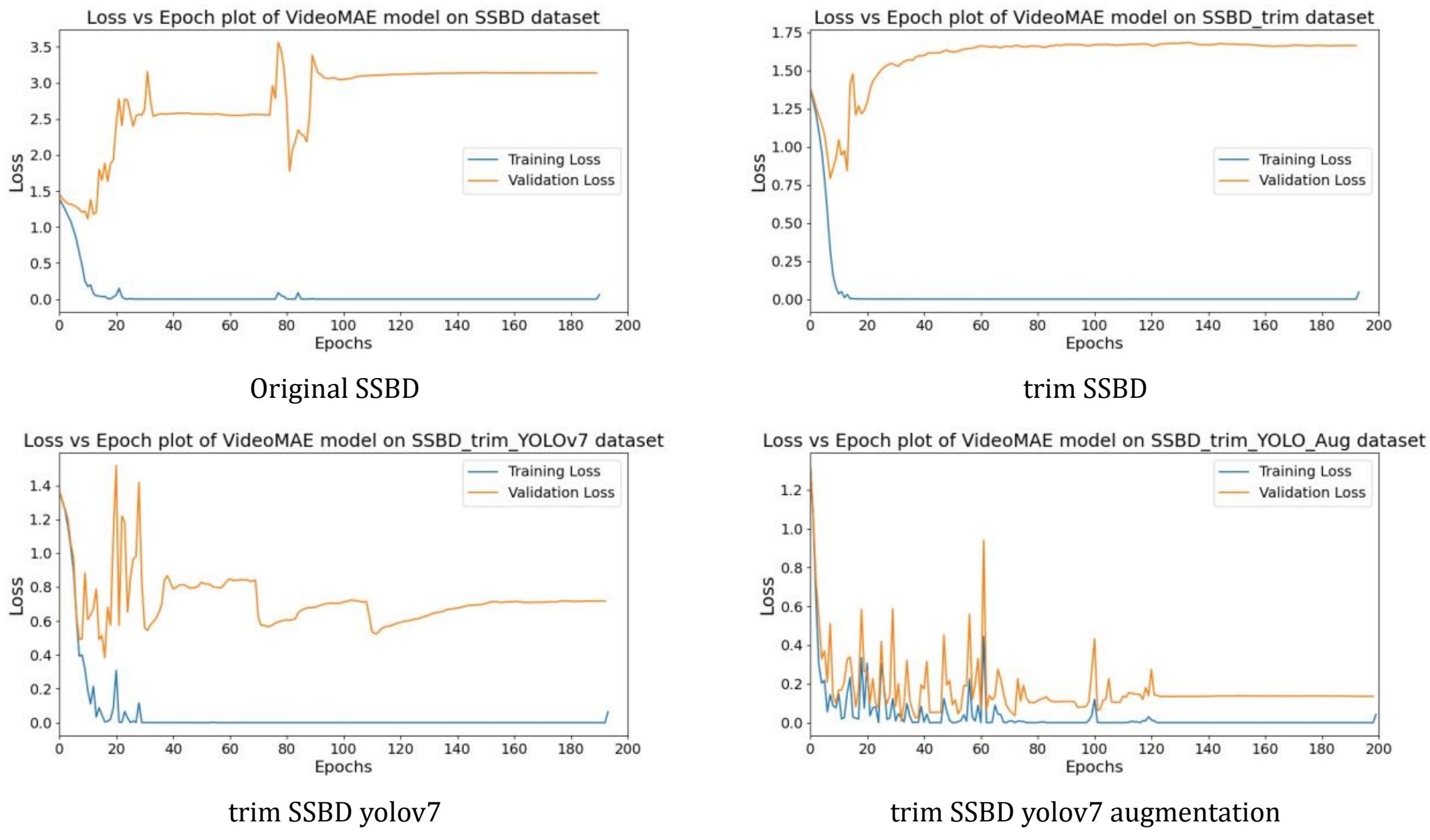

**Fig.18.** Comparison of Loss

The Fig 18 show the training and validation loss trends of the VideoMAE model across different preprocessing methods on the SSBD dataset over 200 epochs. In the Original SSBD graph, training loss rapidly decreases and stabilizes around 0.1, while validation loss fluctuates between 0.5 and 3.5, indicating potential overfitting as the model struggles to generalize well.

In the trim SSBD graph, training loss drops to approximately 0.05, and validation loss stabilizes around 0.1, showing more consistent performance and reduced overfitting. This suggests that trimming enhances the model's ability to generalize by focusing on essential features.

The trim SSBD yolo7 graph displays further reduction in training loss, reaching around 0.03, while validation loss remains stable around 0.05. This indicates improved model stability and learning efficiency, with YOLO-based trimming effectively emphasizing critical features.

In the trim SSBD yolo7 augmentation graph, both training and validation losses are minimized and consistently low, with training loss around 0.02 and validation loss stabilizing at approximately 0.03. This reflects the positive impact of data augmentation in enhancing model robustness and generalization capabilities.

Overall, these graphs demonstrate that strategic preprocessing, especially combining YOLO-based trimming and augmentation, significantly optimizes the VideoMAE model’s learning process by reducing both training and validation losses, thereby improving generalization and performance on the SSBD dataset.

The confusion matrices in fig 19 depicted provide a comparative analysis of the model’s performance in classifying Autism Spectrum Disorder (ASD) and Typically Developing (TD) individuals across four preprocessing variations of the SSBD dataset. In the Original SSBD matrix, the model demonstrates a strong aptitude for identifying ASD, correctly classifying 12 instances with only one misclassified as TD. However, it shows slight confusion in TD prediction, accurately identifying 5 instances but misclassifying 1 as ASD. This initial performance indicates a robust detection capability for ASD but reveals areas for improvement in TD classification accuracy.

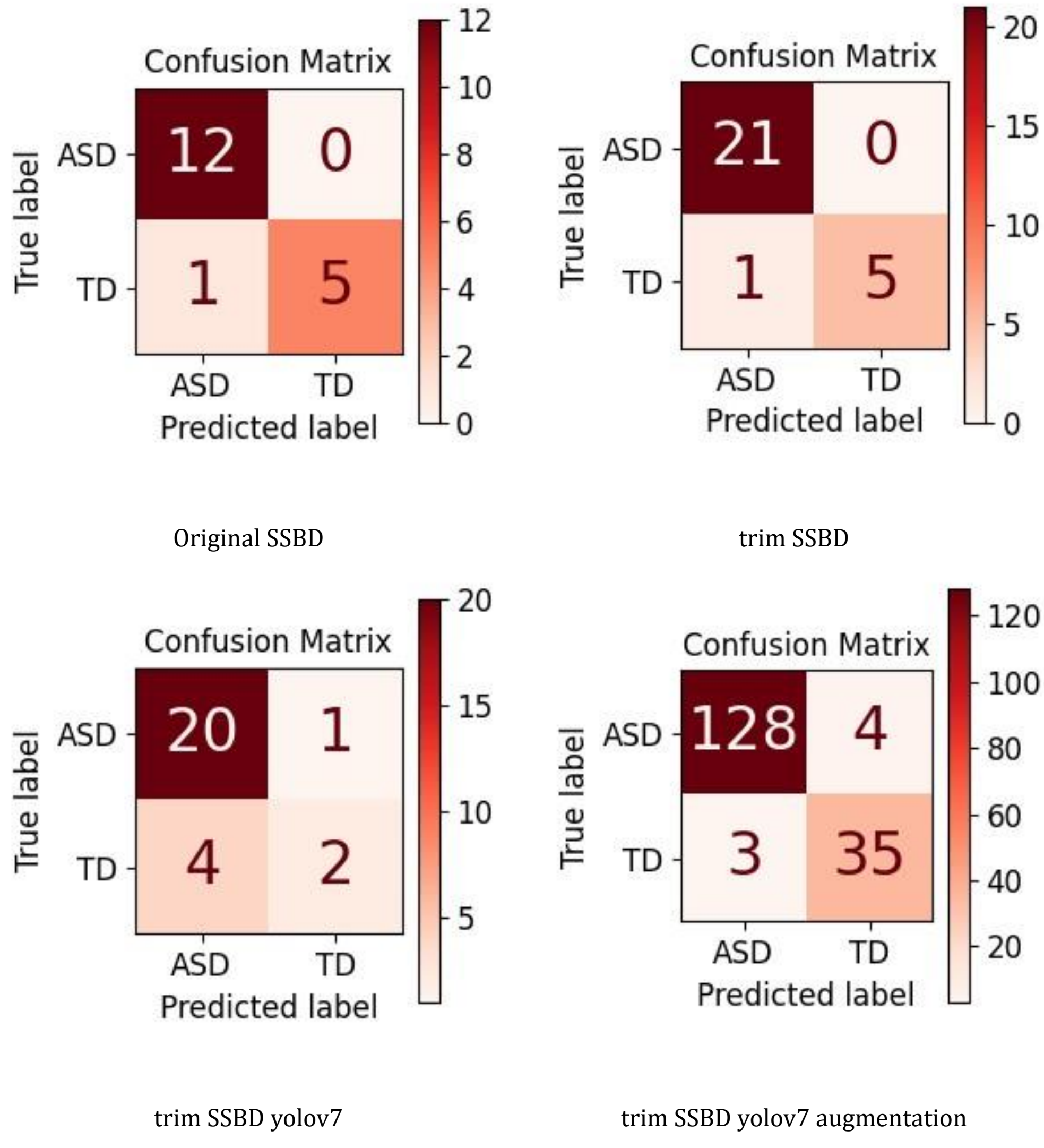

Original SSBD

trim SSBD

trim SSBD yolov7

trim SSBD yolov7 augmentation

**Fig.19.** Comparison of Confusion Matrices

The trim SSBD matrix showcases an enhancement in ASD classification, with 21 correct identifications and only one misclassification. The TD classification remains unchanged, suggesting that the trimming process primarily benefits ASD detection without impacting TD results. This indicates a refined focus on ASD features during preprocessing.

Conversely, the trim SSBD yolo7 matrix illustrates a decline in TD classification accuracy, with only 2 correct identifications and 4 misclassified as ASD, despite maintaining a high accuracy for ASD with 20 correct predictions. This suggests that the YOLO-based trimming may have inadvertently removed critical features necessary for accurate TD classification, leading to increased confusion.

Significant improvements are evident in the trim SSBD yolo7 augmentation matrix, where data augmentation strategies lead to outstanding results. The model accurately classifies 128 ASD instances with only 4 misclassified and achieves 35 correct TD identifications, with 3 misclassified. This highlights the effectiveness of data augmentation in enhancing both recall and precision, substantially improving classification accuracy for both ASD and TD categories. Overall, these matrices underscore the importance of preprocessing techniques, particularly data augmentation, in optimizing model performance and achieving balanced and accurate classification outcomes.

**Table 17.** ASD & TD Results of VideoMAE

| Data Set | Accuracy | Precision | Recall | F1 Score |
|---|---|---|---|---|
| SSBD | 0.94 | 0.96 | 0.91 | 0.93 |
| trim SSBD | 0.96 | 0.97 | 0.91 | 0.94 |
| trim SSBD yolo | 0.81 | 0.75 | 0.64 | 0.66 |
| trim SSBD yolo augment | 0.95 | 0.93 | 0.94 | 0.94 |

The table 17 presents the classification performance of the VideoMAE model across different variations of the SSBD dataset, focusing on metrics such as accuracy, precision, recall, and F1 score. These metrics offer insights into how effectively the model distinguishes between ASD (Autism Spectrum Disorder) and TD (Typically Developing) classes.

For the SSBD dataset, the model achieves a high accuracy of 0.94, indicating a strong rate of correct classifications. Precision is at 0.96, suggesting that the majority of instances identified as positive are indeed true positives. Recall, slightly lower at 0.91, indicates a few true positives were missed. The F1 score, which balances precision and recall, is 0.93, reflecting robust overall performance.

In the trim SSBD dataset, the model's performance improves, with accuracy rising to 0.96. Precision climbs slightly to 0.97, maintaining a strong capability to identify true positives. Recall remains consistent at 0.91, and the F1 score increases to 0.94, highlighting enhanced performance with the trimmed data.

Conversely, the trim SSBD yolo dataset sees a drop in performance. Accuracy falls to 0.81, with precision at 0.75 and recall at 0.64, resulting in an F1 score of 0.66. This suggests that the YOLO-based trimming may have led to the loss of crucial features, impacting the model's classification accuracy negatively.

A notable recovery is evident in the trim SSBD yolo augment dataset, where accuracy rebounds to 0.95. Precision remains strong at 0.93, and recall improves to 0.94, culminating in an F1 score of 0.94. This demonstrates the benefits of data augmentation in restoring and enhancing classification performance, particularly after the performance dip observed with YOLO trimming.

Overall, the table highlights the different impacts of dataset modifications on the VideoMAE model's classification abilities. While certain trimming techniques may pose challenges, data augmentation emerges as a powerful method to improve and sustain high precision, recall, and balanced F1 scores across ASD and TD classifications.

### 4.4 Result Comparison with state of the art

When the proposed model was compared to Lakkapragad et al. [30], the suggested models outperformed the previous models, which input the hand landmarks into an LSTM and MobileNetV2 using MediaPipe to extract them. An F1 score of 84.0 was attained by their model. In this work, Lakkapragad et al. [30]

**Table 18.** A comparative analysis of the proposed model against prior studies on the SSBD dataset, focusing on accuracy and F1-score metrics.

| Model | Method | Results |
|---|---|---|
| Rajagopalan et al. [5] | Bag Of Words (BOW) | 50.7% accuracy |
| Rajagopalan et al. [27] | Histogram of Dominant Motions (HDM) | 76% accuracy |
| Lakkapragad et al. [30] | MobileNetV2, LSTM | F1 score of .84<br>Precision .89<br>Recall .80 |
| Dia et al. [32] | 3D Convnet and Multi-Stage Temporal Convolutional Network | 74.5% accuracy<br>0.73 F1-Score<br>0.73 Recall |
| Ali et al. [33] | Yolov5, DeepSORT, RAFT | 75% accuracy<br>0.90 F1 Score |
| **Our Work** | Yolov7, Augmentations, VideoMAE | **95% accuracy**<br>**.93 precision**<br>**.94 F1 score**<br>**0.94 recall** |

only work with the Arm flapping class. The histogram of dominating motion and Bag of words was combined with the optical flow histogram by Rajagopalan et al. [5,27]. Their classification model distinguished between headbanging, spinning, and hand flapping in three groups with an accuracy of 76.3%. ASD and hand flapping were investigated from all three classes by Ali et al. [33]. Using a large collection of videos depicting the stereotyped behavior of children, Ali et al. [33] investigated the identification of ASD behavior. Their combination of an RGB and optical flow two-stream I3D model produced an accuracy of 75%. Negin F et al. [26] collected a new dataset and recognized autistic and non-autistic children via vision assistance. Multiple models such as LSTM, ConvLSTM, and 3DCNN have been implemented and achieved an accuracy of 79%. 3D CNN and Multi-stage temporal convolution network were implemented by Dia et al. [32] on modified SSBD with all three classes. Dia et al. [32] got 61 videos out of 75 videos from SSBD and collected more videos and generated a performance of 83% accuracy. There is a comparative analysis in Table 18.

We discovered that the VideoMAE models excel primarily due to their innovative use of masked autoencoding, which leverages the inherent redundancy in video data to enhance learning efficiency. By masking a substantial portion of the video frames—up to 90-95%—these models focus on reconstructing the missing content, which compels them to capture the essential temporal and spatial dynamics of the videos. This method of forcing the model to predict large unseen segments encourages a deeper understanding of video content, helping the model learn robust, high-level features that are critical for interpreting complex video sequences. The separation of the encoder and decoder in the VideoMAE architecture allows for processing efficiency, as the encoder handles only the unmasked frames, significantly reducing the computational load. Moreover, this setup minimizes overfitting, as the model must generalize well to perform accurate reconstruction across varied inputs. Such characteristics make VideoMAE particularly effective for self-supervised pre-training on large, unlabeled video datasets, facilitating superior performance on a range of video understanding tasks once fine-tuned.

## 5 Conclusion

In this study, behaviors from videos are analyzed using deep learning and computer vision to classify stereotypical repetitive gestures. We trained our models using videos of children going about their everyday lives in an uncontrolled environment to predict repetitive movements. The videos included three different types of videos: arm flapping, head banging, and spinning. The testing set yielded an 95% score for the VideoMAE model.

We used the VideoMAE model and the insights from our trained models to create a video-based repetitive behavior analysis, which can help parents and psychologists in the future to diagnose aberrant behaviors like arm flapping, head banging, and spinning repetitive gestures which may be an indication of autism. This will increase the study's usefulness and practical applicability. Our objective is to advance the detection of repetitive movements in autistic children by using these contemporary technological advancements, enabling early intervention and subsequent therapy for children.

As a result, this may help autistic children with early detection of gestures and become more adept at social communication, which would lessen the challenges that their families face.

The proposed VideoMAE model provides strong capabilities for detecting repetitive autistic gestures. Repetition in gestures needs to extract spatial as well as temporal features. The VideoMAE model is effective in extracting both spatial and temporal information through its masked autoencoders and attention mechanisms, comparing subsequent frames and detecting repetitive movements in a video. Our findings show that the proposed model is capable of successfully recognizing behavioral patterns in uncontrolled, real-world videos. One of the study's future objectives is the potential for high-quality video collations, which would increase the classification findings' accuracy.

**Acknowledgment**

The authors are grateful to the Ministry of Human Resource and Development and the Indian Institute of Information Technology, Allahabad for supplying the necessary materials required to complete this work.